\documentclass{article}

\PassOptionsToPackage{numbers, compress}{natbib}

\usepackage[preprint]{neurips_2026}

\usepackage[utf8]{inputenc}
\usepackage[T1]{fontenc}
\usepackage{hyperref}
\usepackage{url}
\usepackage{booktabs}
\usepackage{amsfonts}
\usepackage{amsmath}
\usepackage{amssymb}
\usepackage{nicefrac}
\usepackage{microtype}
\usepackage{xcolor}
\usepackage{graphicx}
\usepackage{longtable}
\usepackage{multirow}
\usepackage{array}
\usepackage{textcomp}

\graphicspath{{figures/}}

\newcolumntype{L}[1]{>{\raggedright\arraybackslash}p{#1}}
\newcolumntype{R}[1]{>{\raggedleft\arraybackslash}p{#1}}

\providecommand{\doi}[1]{\href{https://doi.org/#1}{doi:#1}}

\newcommand{\PPARd}{PPAR$\delta$}
\newcommand{\ICfifty}{\ensuremath{\mathrm{IC}_{50}}}
\newcommand{\pKi}{\ensuremath{\mathrm{p}K_\mathrm{i}}}
\newcommand{\pICfifty}{\ensuremath{\mathrm{pIC}_{50}}}
\newcommand{\pECfifty}{\ensuremath{\mathrm{pEC}_{50}}}

\title{From Benchmark to Bench: Can Agents Survive Real-World Drug Discovery?}

\author{%
  \normalfont
  Pierre Llompart\textsuperscript{1} \quad
  Levent Guner\textsuperscript{1} \quad
  Helen Lai\textsuperscript{2} \\
  Alessandro Tibo\textsuperscript{3} \quad
Yijie Xu\textsuperscript{4} \\
  \textsuperscript{1}AMD Silo AI, Stockholm, Sweden \\
  \textsuperscript{2}Molecular AI, Discovery Sciences, R\&D,
  AstraZeneca, Cambridge, UK \\
  \textsuperscript{3}Molecular AI, Discovery Sciences, R\&D,
  AstraZeneca, Gothenburg, Sweden \\
  \textsuperscript{4}AMD Silo AI, Helsinki, Finland
}

\begin{document}

\maketitle

\begin{abstract}
  Agentic systems increasingly coordinate molecular-design tools, but it is
  unclear which layer of the stack limits outcomes on real projects. We
  developed MAGI, an open modular agent that authors objectives, launches and
  monitors optimization, interprets structure--activity relationships, and
  revises its strategy accordingly. MAGI generates molecules either directly
  through the LLM or by delegating to REINVENT 4, with scoring services
  interchangeable behind a common contract. We tested it across nine
  retrospective lead-optimization campaigns from three pharmaceutical
  companies, replayed under fixed temporal cutoffs. Both routes produced valid structures: LLM proposals stayed closer to local
chemistry and reached comparable or higher primary activity in fewer
operations, whereas REINVENT explored broader chemical space. Whether a campaign met its objective depended on the predictive models, not
on the generation route: attainment followed model accuracy on the chemistry
proposed, dropping once that chemistry moved outside the model's
applicability domain. Separately, a
blinded evaluation asked whether the MAGI's output could pass as expert work:
chemists were not able to discriminate agentic proposals from held-out compounds, and
judged the SAR reasoning broadly plausible yet incomplete. Together, these
results position MAGI as a coordination layer pluggable into existing
computational chemistry workflows. The ceiling on real projects, however, remains 
currently set by scorer applicability rather than by tool orchestration.

\end{abstract}

\section{Introduction}
\label{sec:introduction}

Lead optimization advances a compound toward a candidate by balancing potency,
safety and developability across repeated design--make--test--analyze cycles, a
major contributor to drug-development cost \cite{dimasi2016innovation}.
Generative methods such as REINVENT 4 optimize molecules against expert-defined
objectives \cite{loeffler2024reinvent4, brown2019guacamol}, but their scoring
configurations require revision as project evidence and structure--activity
relationships evolve.

Agentic systems promise to manage this process by planning, invoking tools, and
adapting in closed loops
\cite{mbran2024chemcrow, boiko2023coscientist, gottweis2025coscientist}.
Chemistry agents and orchestrators have recently proliferated in literature
\cite{drugagent2024, solovev2025madd, wu2025chematagent, he2026chatinvent,
mozi2026, li2025drugpilot, cidd2025, agentd2025, boltzagents2025, drugr2026,
trace2026, kim2025mtmol, frogent2025, molclaw2026, beyondsmiles2026,
aiagents2026}, yet most evaluations test tool execution or generic molecular
benchmarks rather than scientific performance under real project constraints.
Strong benchmark performance can coexist with elementary reasoning failures and
unreliable confidence
\cite{mirza2025chembench, nezhurina2024alice, williams2024easy}, while
language-model molecular generation alone guarantees neither molecular validity
nor controlled multiparameter optimization \cite{skinnider2024invalid}. An agentic campaign is steered by its scoring services, which act as oracles: every decision the agent makes is expressed through them. Their applicability therefore bounds what the agent can achieve. Systematic bias shifts the numerical scale, range compression suppresses real differences, and poor rank correlation makes candidate ordering unreliable. Across the nine campaigns reported here, attainment tracks the applicability of models and success from generation route.

MAGI is an open, modular agent that authors and adapts project objectives,
launches and monitors optimization campaigns, and interprets emerging SAR to
redirect them. Molecular generation occurs either directly through the LLM or
through agent-controlled REINVENT 4, with both routes using a common prediction
and structure-evaluation stack. We evaluate MAGI on four GSK series
\cite{gsk046series, nmethylpyridineseries, messenger2021template, taylor2021del}, four Sanofi projects 
\cite{mas2026retrodmta}, together with one undisclosed AstraZeneca 
 campaign, all retrospectively replayed under fixed temporal cutoffs. Our contributions are:

\begin{itemize}
  \item An auditable modular agent architecture in which project objectives are
        executable, versioned, and shared across LLM and REINVENT generation
        routes.
  \item A first-time evaluation on nine industry campaigns from three pharmaceutical companies 
        reporting validity, objective attainment, workload, and limitations.
  \item An evidence that objective attainment on drug discovery projects is bound by the reliability of predictive models rather than by agent coordination.
\end{itemize}

\section{Related works}
\label{sec:related}

REINVENT 4 is a widely used production framework for goal-directed molecular
generation and optimization \cite{loeffler2024reinvent4}. It maximizes an
expert-defined scoring function but does not author or revise that objective,
making it a natural generator within an agentic workflow. LLM molecular design
systems differ in where generation and critique occur. MT-Mol generates SMILES
through specialized analysis, design, and verification agents
\cite{kim2025mtmol}, whereas CIDD combines an LLM medicinal-chemistry critic
with pocket-conditioned 3D generators \cite{cidd2025}. DrugR uses explicit
pharmacological reasoning \cite{drugr2026}; MolClaw combines REINVENT 4, Boltz-2
and interaction profiling \cite{molclaw2026}; Mozi uses governed agents for
molecular optimization \cite{mozi2026}; and ChatInvent integrates REINVENT and
synthesis planning in a production-facing interface \cite{he2026chatinvent}.
Validation on temporally held-out project chemistry remains rare. RetroDMTA
provides four time-resolved industrial programs \cite{mas2026retrodmta}, but its
original benchmark prioritizes known future compounds. MAGI instead exposes only
pre-cutoff context during generation and uses future compounds for retrospective
comparison.

\section{Method}
\label{sec:method}

\subsection{System}
\label{sec:system}

MAGI separates agent orchestration, molecular generation, and evaluation.
Claude Sonnet 4.6 accesses REINVENT 4 \cite{loeffler2024reinvent4}, global
ADMET models, temporally restricted project models \cite{mas2026retrodmta},
Boltz-2 \cite{passaro2025boltz2}, PoseBusters
\cite{buttenschoen2024posebusters}, and an interaction profiler through a
common REST contract exposed by the Model Context Protocol
\cite{anthropic2024mcp}. Project objectives are represented as versioned
therapeutic target profiles (TTPs) that compile into executable REINVENT
scoring configurations. The MCP server exposes 24 tools covering data access,
TTP management, scoring, campaign control, and result retrieval. Services,
target constructs, and the architecture are documented in
Table~\ref{tab:services}, Table~\ref{tab:constructs}, and
Figure~\ref{fig:architecture}a.

Optimization jobs execute asynchronously and are monitored by the gateway.
Independently containerized services exchange structures, configurations, and
results through shared storage. An external REINVENT component batches
generated SMILES to these services, while a dependency-aware wrapper passes
Boltz-2 complexes to PoseBusters and the interaction profiler. Structural
evidence can therefore contribute directly to optimization rather than only
post hoc ranking.

\subsection{Objective and configuration space}
\label{sec:objective}

Each campaign is governed by a versioned, machine-readable therapeutic target
profile. Each objective specifies its scoring service, optimization
direction, target value or acceptable interval, desirability transform, and
weight; fixed guardrails are evaluated separately as pass/fail constraints.
Endpoint predictions are transformed to $[0,1]$ desirabilities and combined
using a weighted Derringer--Suich geometric mean
\cite{derringer1980}. The complete formulation is given in
Section~\ref{sec:ttp-scoring}. The same TTP evaluates LLM proposals and compiles
the REINVENT scoring configuration; every agent-authored revision is versioned,
and missing values remain unevaluable. Project-specific objectives and
guardrails are reported in Table~\ref{tab:objectives}.

\section{Experimental setup}
\label{sec:setup}

\paragraph{Datasets.}
MAGI is evaluated on nine lead-optimization tasks: four published GSK BET BD2
series \cite{gsk046series, nmethylpyridineseries, messenger2021template,
taylor2021del}, four Sanofi RetroDMTA projects---FXA, MMP8, \PPARd{} and
RENIN \cite{mas2026retrodmta}---and one undisclosed AstraZeneca campaign. Each
task shows the agent one starter and four measured analogues and withholds every
remaining compounds. The primary endpoints are kinetic solubility for GSK-S1, membrane permeability
for GSK-S2 and $\log D_{7.4}$ for GSK-S3 and GSK-S4; $\log D_{7.4}$
for AstraZeneca; \pKi{} for FXA, \pICfifty{} for MMP8 and RENIN, and \pECfifty{} for \PPARd{}, with
solubility and $\log D_{7.4}$ as additional FXA objectives. Sanofi project models see only pre-cutoff compounds, so the
future cohort remains held out at both generation and scoring. Endpoint
definitions, cohort sizes and splits are reported in
Tables~\ref{tab:admet-endpoints} and~\ref{tab:endpoint-composition}.

\paragraph{Protocol.}
Every campaign ran for five iterative optimization rounds. Each round supplies
the project context and the current objective, generates and evaluates
molecules, and ends with an agent-authored shortlist of approximately five
compounds carried into the next round. On the REINVENT route the agent
additionally creates or updates a TTP, launches and monitors optimization, and
retrieves the generated pool from which it selects those compounds.
Figure~\ref{fig:architecture}b summarizes the protocol, and
Tables~\ref{tab:pool} and~\ref{tab:reinvent-params} report pool sizes, shortlist
construction, cofolding coverage, and generation parameters.

\paragraph{Metrics.}
Molecular quality is summarized by validity, cumulative uniqueness, novelty to
the shown set, similarity to held-out chemistry, intra-round diversity and
synthetic accessibility \cite{ertl2009sascore, rogers2010ecfp}, where diversity
is one minus the mean pairwise ECFP4 Tanimoto similarity. Cofolding is
summarized by Boltz-2 ligand iPTM and interaction counts. Operational load is
the number of observed generation, prediction, cofolding, validation,
monitoring, retrieval and inspection calls. Project- and route-level results
are reported in Figure~\ref{fig:main-quality} and
Figures~\ref{fig:round-similarity}--\ref{fig:renin-mmp8-properties}.

\section{Results}
\label{sec:results}

\subsection{Molecular quality and chemical-space exploration}
\label{sec:quality}

We first assess the validity and uniqueness of proposals from each generative route.
Figure~\ref{fig:main-quality}a compares round-level validity, cumulative
uniqueness, novelty to shown chemistry, similarity to held-out chemistry and
intra-round diversity across the GSK and Sanofi projects. Route differences
depended on dataset and objective. GSK proposals were more novel overall, and
within Sanofi it was REINVENT that departed furthest from the shown chemistry;
conversely, Sanofi proposals resembled the held-out compounds more closely than
GSK ones did. REINVENT achieved greater intra-round diversity in both
organizations, supporting broader exploration.

\subsection{Primary-endpoint and objective-attainment evaluation}
\label{sec:attainment}

\begin{figure}[!t]
  \centering
  \includegraphics[width=\linewidth]{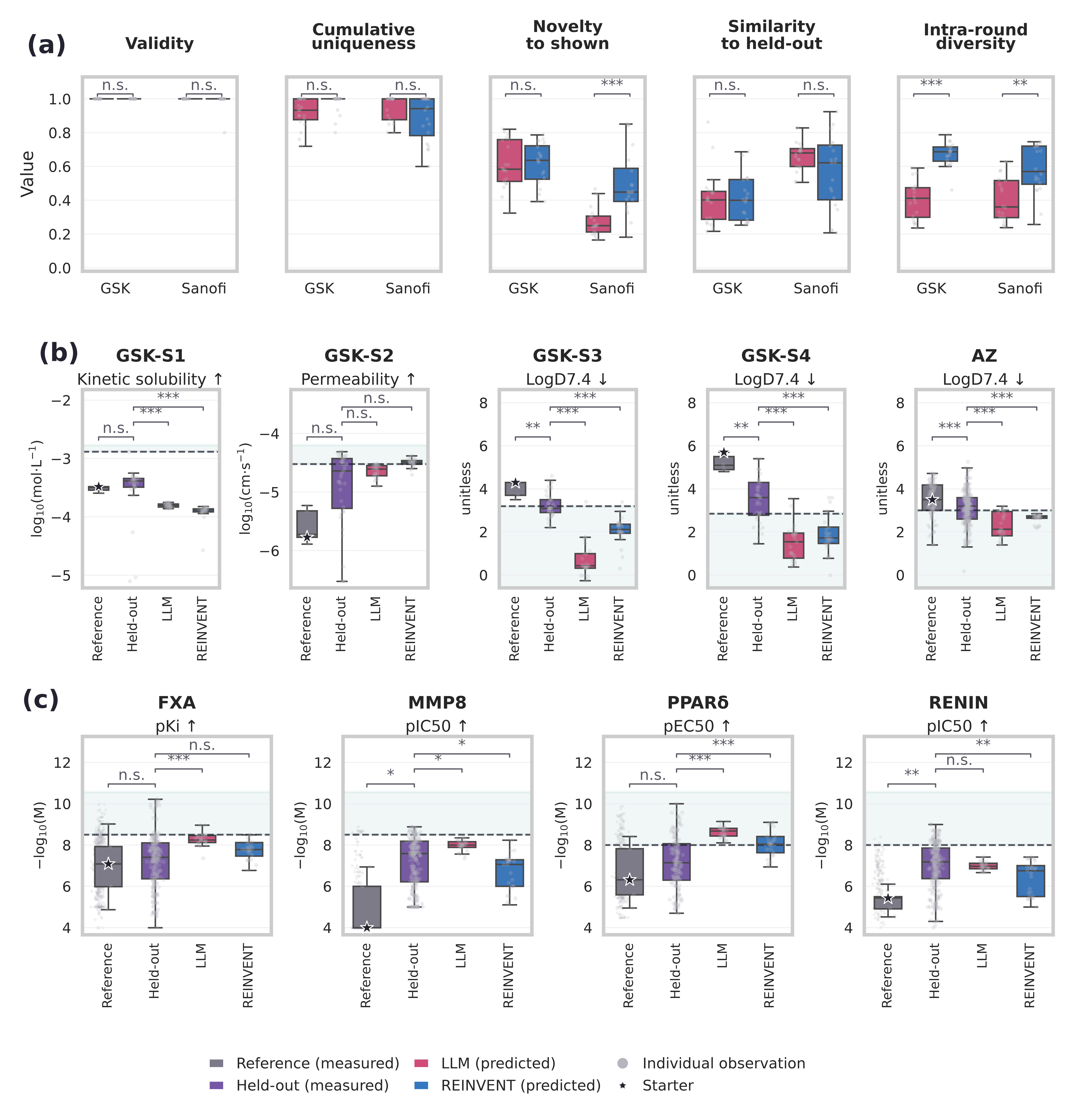}
  \caption{\textbf{Molecular quality and objective attainment across Drug Discovery campaigns.} (a) Project-round validity,
    cumulative uniqueness, novelty, held-out similarity, and intra-round
    diversity for LLM (red) and REINVENT (blue). (b) GSK and AstraZeneca primary-endpoint distributions. (c) Sanofi project-level
    primary-endpoint distributions. Pale-green shading indicates the
    objective-satisfying region. Pale points denote pre-cutoff Reference compounds. Stars mark the selected starter compound. Brackets show
    Holm-adjusted two-sided Mann--Whitney U tests, adjusted p-value ($p$)
    significance ($*$ $p < 0.05$, $**$ $p < 0.01$, $***$ $p < 0.001$; n.s., not
    significant).}
  \label{fig:main-quality}
\end{figure}

Chemical validity alone does not establish whether a campaign advances its
project objective. We therefore compare each route's proposals against two
experimental references on the primary endpoint: the analogues shown to the
agent, and the held-out compounds it never saw
(Figure~\ref{fig:main-quality}b for GSK and AstraZeneca,
Figure~\ref{fig:main-quality}c for Sanofi).

Three of the four GSK series and the AstraZeneca campaign met their objective.
GSK-S1 was the exception: its kinetic-solubility target sits outside the training range  (Table~\ref{tab:oof-performance}). During the experiment, the agent autonomously promoted thermodynamic solubility as a proxy, followed by an improvement in proposals (Figure~\ref{fig:gsk12-properties}). On the Sanofi campaigns, both
routes matched the held-out compounds on predicted primary activity, yet only
FXA and \PPARd{} cleared the project threshold. The LLM reached the objective more often
than REINVENT, reproducing the quality these programmes historically
reached without surpassing it. In case where a
campaign fell short, the cause coincided with a model operating
outside its applicablity range: GSK-S1
failed because its scorer was extrapolating, while
RENIN ligand-based model showed weak held-out correlation
(Figure~\ref{fig:model-validation}). Supporting analyses report round-wise diversity
(Figure~\ref{fig:round-similarity}), synthetic accessibility and chemical-space
 (Figures~\ref{fig:structural-quality}--\ref{fig:umap-sanofi}),
models validation (Figure~\ref{fig:model-validation}), trajectories
(Figures~\ref{fig:gsk12-properties}--\ref{fig:renin-mmp8-properties}) and molecules
(Figures~\ref{fig:gsk-best}--\ref{fig:sanofi-top5}).

\subsection{Operational behaviour and medicinal chemistry evaluation}
\label{sec:operations}

\begin{figure}[t]
  \centering
  \includegraphics[width=\linewidth]{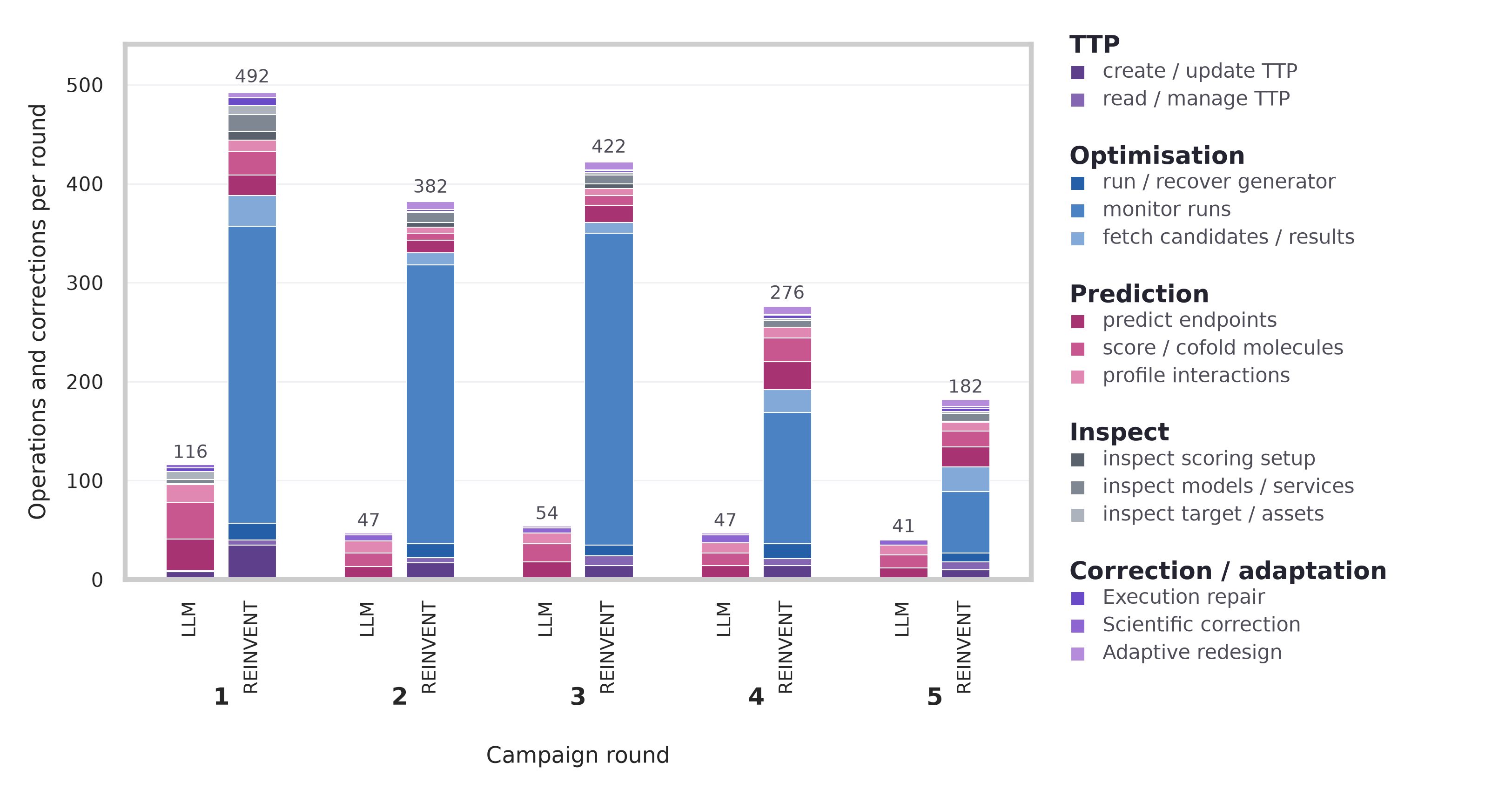}
  \caption{\textbf{Observable agent operations and correction/adaptation
    events.} Counts are shown by category, campaign round, and generation route;
    counts exclude scorer and model calls executed internally within REINVENT's
    reinforcement-learning loop.}
  \label{fig:operations}
\end{figure}

The routes produced similar profiles but set different
operational demands. Figure~\ref{fig:operations} summarizes  
operations and adaptation events. REINVENT required four-to-nine times more operations than the LLM (Figure~\ref{fig:tokens}), arising
from launch, monitoring, recovery, and retrieval. These
counts exclude the calls issued inside the reinforcement-learning loop:
each REINVENT campaign explored 100,000 more structures than the LLM, increasing the number of tool calls to four more orders of
magnitude. MAGI's adaptations extended beyond syntax repair: it revised
objectives, search parameters, and molecular strategy in response to
convergence, scorer disagreement, and emerging SAR
(Section~\ref{sec:adaptation}). In a blinded evaluation involving four chemists and 28 questions, participants
were unable to distinguish AI-generated from held-out molecules, showed no
preference between design routes, and found the agent's
SAR reasoning plausible but incomplete
(Section~\ref{sec:medchem-eval}, Figure~\ref{fig:medchem-evaluation}).

\section{Discussion}
\label{sec:discussion}

The two generative routes expose an exploration--exploitation trade-off. LLM
generation acted as a focused local optimizer around the shown chemistry,
whereas agent-controlled REINVENT broadened diversity and novelty when a project
needed to leave its starting chemotype. Neither route consistently dominated
model-space attainment, indicating that generator breadth cannot compensate for
a biased, range-compressed, or weakly ranking scorer. Structure-based services
shifted proposal rankings in a target-dependent manner: reranking by Boltz-2
affinity changed the top-5 shortlist substantially 
(Figures~\ref{fig:heldout-similarity} and~\ref{fig:model-validation}d). The
blinded evaluation highlighted that chemists could not separate
agent proposals from measured compounds nor that they a preference for held-out over AI compounds. MAGI's contribution lies
above the orchestration, in selecting which tool to invoke and revising the
objective as evidence accumulates \cite{kim2025mtmol,
beyondsmiles2026}.

\section{Limitations}
\label{sec:limitations}

Proposal endpoints are predictions, not measurements. Each condition
used one backend, one trajectory, and no static-objective control, so the
agent's own contribution to attainment is not isolated. Temporal splits
restrict train/test overlap but cannot exclude pretraining exposure.

\section{Conclusion}
\label{sec:conclusion}

We presented MAGI, a modular reasoning layer that versions project objectives,
coordinates established generators and predictors, monitors runs, and
adapts execution as evidence accumulates. Across nine industrial campaigns,
LLM generation provided local optimization with fewer operations, while
REINVENT broadened exploration at higher operational cost.
Neither route consistently won on objective attainment as temporal controls
show this depended on the scoring models: campaigns succeeded when predictors were reliable and
failed otherwise. Real project data surfaced this
limit in a way benchmarks would not, from weak model performance on industrial 
data to autonomous adaptation by the agent through proxy endpoints. Overall, agentic design survives contact
with real projects, but two critical ceilings remain: the quality of the agent's SAR
reasoning, and the applicability of scoring services in an industrial context.

\medskip
{
\small
\bibliographystyle{unsrtnat}
\bibliography{references}
}

%%%%%%%%%%%%%%%%%%%%%%%%%%%%%%%%%%%%%%%%%%%%%%%%%%%%%%%%%%%%

% Flush every pending main-text float before the counters are reset, so a
% deferred Figure 2 cannot be renumbered into the supplementary sequence.
\clearpage

\appendix

% The supplementary material uses S-prefixed numbering (Section S1,
% Figure S1, Table S1) rather than the \appendix default of A/B/C, matching
% every cross-reference inherited from the source manuscript.
\setcounter{section}{0}
\setcounter{figure}{0}
\setcounter{table}{0}
\renewcommand{\thesection}{S\arabic{section}}
\renewcommand{\thesubsection}{S\arabic{section}.\arabic{subsection}}
\renewcommand{\thefigure}{S\arabic{figure}}
\renewcommand{\thetable}{S\arabic{table}}
\renewcommand{\theequation}{S\arabic{equation}}

% hyperref builds anchor names from \theH<counter>, which still uses the reset
% counter values. Without these, Figure S1 collides with Figure 1.
\renewcommand{\theHsection}{S\arabic{section}}
\renewcommand{\theHsubsection}{S\arabic{section}.\arabic{subsection}}
\renewcommand{\theHfigure}{S\arabic{figure}}
\renewcommand{\theHtable}{S\arabic{table}}
\renewcommand{\theHequation}{S\arabic{equation}}

\section*{Supplementary Information}
\addcontentsline{toc}{section}{Supplementary Information}

\section{Extended related work}
\label{sec:extended-related}

Agentic systems for chemistry began with tool-augmented assistants: ChemCrow
coupled a language model with chemistry tools \cite{mbran2024chemcrow},
Coscientist further leveraged wet-lab automation to drive experiments
\cite{boiko2023coscientist}, and multi-agent architectures refined hypotheses
\cite{gottweis2025coscientist}. Drug-discovery orchestrators followed, with
pipeline assembly from natural language \cite{drugagent2024, solovev2025madd},
tool selection by tree search \cite{wu2025chematagent}, production deployment
around REINVENT \cite{he2026chatinvent}, governed hierarchies \cite{mozi2026},
parameterized memory \cite{li2025drugpilot}, a critic over a structure-based
generator \cite{cidd2025}, modular task execution \cite{agentd2025} and agentic
affinity prediction \cite{boltzagents2025}. Agents aimed specifically at lead
optimization add explicit-reasoning optimization \cite{drugr2026},
trajectory-aware tool planning \cite{trace2026}, role-specialized multi-agent
optimization \cite{kim2025mtmol}, and a full-process system on the Model Context
Protocol \cite{frogent2025, anthropic2024mcp}. MolClaw is the closest published
stack to ours \cite{molclaw2026, beyondsmiles2026, aiagents2026}.

\section{Agent}
\label{sec:agent}

\subsection{System}
\label{sec:agent-system}

The gateway agent runs a bounded tool-use loop (at most 100 tool rounds per
agent turn) and writes observable model messages, tool requests, and tool
results to a structured JSONL transcript; a background monitor polls job health
and injects explicit alerts. The therapeutic target profile (TTP) is a
schema-validated, versioned JSON object. Each native or external objective
identifies its scoring service, optimization direction, target value or
acceptable interval; desirability transforms and, when applicable, weight.
Dependencies and aggregation rules define the composite score, while fixed
guardrails and SMARTS-based structural constraints are evaluated separately as
pass/fail criteria. Each campaign starts with TTP version 1 before the first
round. The primary endpoint, data split or temporal cutoff, and fixed guardrails
remain unchanged throughout the campaign. Secondary objectives, weights, or
search settings may change only through a recorded TTP or configuration update,
which increments the stored version. A mapper topologically orders objective
dependencies and compiles the TTP into an executable REINVENT scorer
configuration \cite{loeffler2024reinvent4}. The same TTP semantics are used to
evaluate molecules proposed directly by the LLM and molecules generated by
REINVENT, ensuring that both routes are compared under the same objectives and
guardrails. Each service exposes a POST /score endpoint used by the TTP and by REINVENT
scoring, and, where implemented, a POST /predict endpoint that the agent can
call directly for analysis. Boltz-2 returns
cofolding, affinity, confidence, and structure artifacts
\cite{passaro2025boltz2}; PoseBusters consumes the corresponding upstream
complex \cite{buttenschoen2024posebusters}; the general ADMET and RetroDMTA
services return endpoint predictions and uncertainty metadata
\cite{mas2026retrodmta}. Tool exceptions are returned to the agent as errors.
External REST scoring failures are logged and represented as NaN unless an
explicit failure score is configured. Missing service outputs remain unevaluable
and are not counted as objective attainment; a language-model estimate is never
substituted for a missing service value. The architecture separates decisions
from evidence generation: the gateway chooses a route and manages a versioned
objective, while molecular generation, prediction and structural checks remain
independently replaceable as services.

\begin{figure}[htbp]
  \centering
  \includegraphics[width=\linewidth]{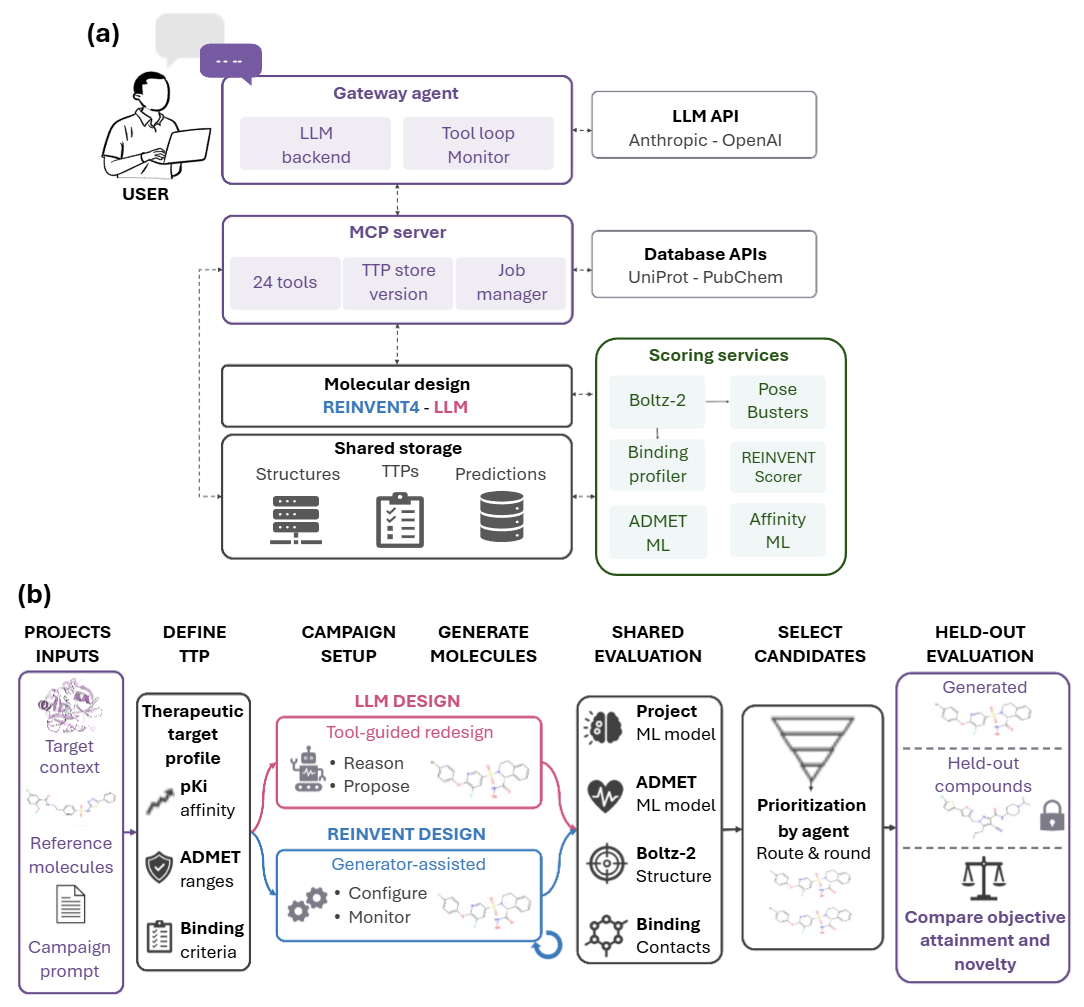}
  \caption{\textbf{MAGI service interaction, MCP interface, and experimental
    protocol.} \textbf{(a) System architecture.} The gateway agent connects the
    LLM reasoning backend to a monitored tool loop. Through the MCP server, it
    manages versioned therapeutic target profiles, optimization jobs, REINVENT
    4, external data sources, and independently deployable scoring services.
    Shared storage transfers structures, configurations, and results between
    services without placing large artifacts in the language-model context. A
    TTP defines the scoring pipeline; molecules proposed by the LLM or generated
    by REINVENT are evaluated in batches using native molecular properties,
    Boltz-2 cofolding, pose and interaction analysis, global ADMET models and
    project-specific RetroDMTA models. Their component outputs are aggregated
    and returned to the generation loop to guide the next iteration. The MCP
    server exposes 24 tools across six groups: service discovery, data
    retrieval, TTP management, scoring and prediction, optimization control, and
    result retrieval. \textbf{(b) Controlled campaign protocol.} Project
    context, reference molecules, and objectives are encoded in a versioned TTP.
    LLM and REINVENT generate candidates through distinct routes but use the
    same activity, ADMET, Boltz-2, and interaction-scoring services. Ranked
    evidence guides subsequent rounds, after which final candidates are compared
    with measured held-out compounds never shown to the agents.}
  \label{fig:architecture}
\end{figure}

\subsection{Services usage}
\label{sec:services-usage}

Both generation routes access the same molecular-evaluation services and author
a TTP to score the molecules they produce. The LLM route asks the reasoning
backend to propose a shortlist directly; the REINVENT route additionally
optimizes the TTP, monitors the job, retrieves the generated pool, and selects
the final shortlist from it.

Which services a campaign uses depends on the organization. GSK and AstraZeneca
campaigns used the general ADMET stack and excluded cofolding, whereas Sanofi
campaigns used the target-specific project model and required
Boltz-2/PoseBusters structural evaluation. Boltz-2 produced a complex and an
affinity prediction, and PoseBusters consumed that exact complex for geometry evaluation. Every service
in Table~\ref{tab:services} was reachable either inside the executable TTP,
through the common POST /score contract, or directly by the agent for analysis;
the general ADMET and project-specific services support both modes.
Table~\ref{tab:services} reports the deployed versions, outputs, and quality
controls. Prediction requests were batch-oriented and preserved input order.

\begin{table}[htbp]
  \caption{Agent-accessible services, model versions, outputs, roles, and
    applicability or quality controls.(MTL: Multi-Task Learning)}
  \label{tab:services}
  \centering
  \footnotesize
  \setlength{\tabcolsep}{3pt}
  \begin{tabular}{@{}L{0.11\linewidth}L{0.17\linewidth}L{0.22\linewidth}L{0.17\linewidth}L{0.24\linewidth}@{}}
    \toprule
    \textbf{Service} & \textbf{Model/version} & \textbf{Endpoints or outputs} &
    \textbf{Role} & \textbf{Applicability or quality controls} \\
    \midrule
    \textbf{REINVENT 4} & REINVENT 4.8.24; Mol2Mol &
    Generated SMILES, component/aggregate reward &
    REINVENT molecular generation &
    Schema-validated TTP; RDKit pre-filters; diversity filter; agent-selected
    seeds and hyperparameters \\
    \addlinespace
    \textbf{General ADMET predictor} &
    Consensus of XGBoost, Chemprop-MTL and CheMeleon-MTL &
    logP, $\log D_{7.4}$, thermodynamcic/kinetic solubility, Caco-2 permeability,
    MDCK/Caco-2 efflux &
    ADMET endpoint scoring and optional secondary objectives &
    Applicability domain, Tanimoto similarity to proposal \\
    \addlinespace
    \textbf{RetroDMTA predictor} &
    Target/window ensemble of XGBoost, Chemprop-MTL and CheMeleon-MTL &
    FXA \pKi{}/logS/$\log D_{7.4}$; MMP8 \pICfifty{}; \PPARd{} \pECfifty{};
    RENIN \pICfifty{} &
    Project-specific ligand scoring &
    Target/window locked per campaign; cross-model SD \\
    \addlinespace
    \textbf{Boltz-2} & Boltz 2.2.1 &
    Predicted complex, predicted affinity and ligand iPTM &
    Cofolding and structural co-objective &
    RDKit ligand QC; supplied MSA; standard amino acids; $\leq$56 heavy and
    $\leq$128 total atoms \\
    \addlinespace
    \textbf{PoseBusters} & PoseBusters 0.6.5 &
    Per-check booleans and fraction of checks passed &
    Physical-validity assessment of Boltz poses &
    Invalid/missing complexes return null \\
    \addlinespace
    \textbf{Interaction profiler} &
    RDKit, OpenMM-based service &
    Typed contacts, interaction counts and fingerprints &
    Binding-mode interpretation &
    SMILES/structure validation; ligand-identity check; 8~\AA{} radius \\
    \bottomrule
  \end{tabular}
\end{table}

\subsection{TTP scoring semantics}
\label{sec:ttp-scoring}

For objective $i$, the TTP records a scoring service, optimization direction,
target value or acceptable interval, desirability transform $T_i$, and weight
$w_i$. For molecule $m$, the service prediction $x_i(m)$ is transformed to

\begin{equation}
  d_i(m) = T_i\bigl(x_i(m)\bigr) \in [0,1].
  \label{eq:desirability}
\end{equation}

Endpoint desirabilities are combined using the weighted Derringer--Suich
geometric mean:

\begin{equation}
  D(m) =
  \left(
    \prod_i d_i(m)^{w_i}
  \right)^{1/\sum_i w_i}.
  \label{eq:geometric-mean}
\end{equation}

Here, $D(m)$ is the overall desirability and $1-D(m)$ the distance to the
profile. The geometric mean prevents strong performance on one endpoint from
compensating for complete failure on another \cite{derringer1980}. Complete
TTP attainment is reported separately and requires every mandatory objective
and fixed guardrail to pass. Missing service values remain unevaluable and are
not treated as attainment.

\subsection{Scoring contract}
\label{sec:scoring-contract}

External scorers use an index-aligned REST contract: POST/score gets a SMILES
list and returns scores, metadata, and artifacts. Agent-facing POST /predict
endpoints provide analysis without requiring a TTP. The MCP scoring layer
formats each molecule with total and component scores, raw and transformed
values, failure reasons, warnings, TTP identity/version, and available structure
paths. Pipeline dependencies are passed by artifact identity, so PoseBusters
consumes the exact Boltz-2 complex associated with each molecule. Network or
HTTP failures in the external scorer are logged and produce NaN by default; an
explicit native-only fallback is labelled as such.

\subsection{System prompt}
\label{sec:system-prompt}

The agent system prompt defines the agent as a medicinal-chemistry
orchestration layer, requires tool-grounded molecular claims, and specifies TTP
authoring, monitoring, selection and terminal reporting. Per-round task prompts
provide project chemistry, objectives, guardrails, and route-specific
constraints. The configured backend was Claude Sonnet 4.6; provider-side
sampling defaults were used, and no random seed was fixed for the reasoning
model. We analyse only observable messages, tool requests, tool results, and
configuration payloads.

\subsection{Agent adaptation, error recovery, and TTP provenance}
\label{sec:adaptation}

For every campaign, MAGI records each observable model message, tool request,
tool result, TTP version, and executable scoring configuration. Scientific
adaptations are evidence-motivated changes to objectives, weights,
transformations, search settings, or design strategy and are recorded separately
from execution of repairs. For example, if candidates repeatedly fail a
solubility guardrail and the agent consequently increases the solubility weight,
the guardrail failures are the preceding evidence, and the weight change is the
scientific adaptation. A change is classified as a scientific adaptation only
when the transcript explicitly links preceding campaign evidence to the
subsequent change. Syntax repairs, routine initialization, status polling, and
repeated serialization are classified separately as execution work. An
operational error is an observable failed or invalid tool interaction, including
service unavailability, malformed input, schema-validation failure, job failure,
missing artifact, timeout, or failed results retrieval. A recovery is counted
when a subsequent observable agent action resolves the error, and the intended
operation completes successfully. Recovery is classified as autonomous when no
human intervention or manual configuration change occurs between the error and
successful completion. Repeated status polling without an error is not counted
as recovery.

\subsection{Compute requirements}
\label{sec:compute}

Experiments ran in containerized services on nodes equipped with AMD EPYC 9575F
CPUs and AMD Instinct MI355X accelerators under ROCm 7.0. REINVENT and Boltz-2
used separate GPUs; structure standardization, PoseBusters and tabular
prediction were CPU-parallelized. Tables~\ref{tab:pool}
and~\ref{tab:reinvent-params} report generation workloads and
Figure~\ref{fig:tokens} reports input-token volume versus observable operations.

\section{Datasets}
\label{sec:datasets}

\subsection{Global ADMET data}
\label{sec:admet-data}

Raw experimental records were obtained from OCHEM \cite{sushko2011ochem} and
ChEMBL \cite{zdrazil2024chembl}. Assay metadata was harmonized before combining
sources. $\log D_{7.4}$ records included chromatographic $\log D_{7.4}$ measurements. Agreement between project-specific $\log D_{7.4}$ and global ADMET $\log D_{7.4}$ is
verified in Figure~\ref{fig:model-validation}a, with the $\log D_{7.4}$ model
demonstrating high accuracy over the GSK project data. Solubility records were
restricted to aqueous-buffer measurements at pH 7.4 and separated into
thermodynamic aqueous solubility and kinetic solubility measured after dilution
of a DMSO stock into buffer. Kinetic-solubility records included measurements
obtained after dilution of a DMSO stock into aqueous buffer, including
CLND-reported (ChemiLuminescent Nitrogen Detection) measurements when they
satisfied the same protocol and unit-curation criteria. Caco-2 permeability
retained inhibitor-free apical-to-basolateral (A$\rightarrow$B) measurements.
Efflux measurements were also required to be inhibitor-free and were retained as
separate Caco-2 and MDR1--MDCK endpoints. Transport assays used compound
concentrations below 100~\textmu{}M.

After the structure, unit and replicate curation described in
Section~\ref{sec:data-processing}, the multitask dataset contained 156,797
unique molecules with one or more labels across seven endpoints: logP, $\log D_{7.4}$,
aqueous thermodynamic solubility, kinetic solubility, Caco-2 permeability, MDR1--MDCK efflux ratio and Caco-2 efflux ratio. The sparse multitask dataset contains 156,797 unique
molecules and 186,120 datapoints.

\begin{table}[htbp]
  \caption{Global ADMET endpoint definitions, curated datapoint counts, and
    retained value ranges.}
  \label{tab:admet-endpoints}
  \centering
  \small
  \begin{tabular}{@{}L{0.24\linewidth}L{0.32\linewidth}rrr@{}}
    \toprule
    \textbf{Endpoint} & \textbf{Conditions} & \textbf{Datapoints} &
    \textbf{Minimum} & \textbf{Maximum} \\
    \midrule
    LogP & -- & 56,140 & $-3.0$ & 7.0 \\
    \addlinespace
    $\log D_{7.4}$ & PBS Buffer pH 7.4 \newline Chromatography &
    33,628 & $-3.0$ & 7.0 \\
    \addlinespace
    Thermodynamic solubility &
    PBS Buffer $7.4\pm0.5$ \newline Temperature $25\pm5\,^\circ$C \newline
    No excipients or cosolvents &
    32,078 & $-8.0$ & 1.0 \\
    \addlinespace
    Kinetic solubility & Temperature $25\pm5\,^\circ$C &
    48,116 & $-6.5$ & $-3.7$ \\
    \addlinespace
    Permeability Caco-2 &
    No engineered cell lines \newline No P-gp inhibitors \newline
    Below 100~\textmu{}M concentration \newline Below 20 cell passages \newline
    Apical to basolateral transport &
    9,600 & $-8.0$ & $-4.0$ \\
    \addlinespace
    Efflux ratio Caco-2 &
    No engineered cell lines \newline No P-gp inhibitors \newline
    Below 100~\textmu{}M concentration \newline Below 20 cell passages &
    2,007 & $-0.5$ & 2.0 \\
    \addlinespace
    Efflux ratio MDCK-MDR1 &
    No engineered cell lines \newline No P-gp inhibitors \newline
    Below 100~\textmu{}M concentration \newline Below 20 cell passages &
    4,551 & $-0.5$ & 2.9 \\
    \bottomrule
  \end{tabular}
\end{table}

\subsection{BET BD2 ADMET series}
\label{sec:bet-series}

The four series were extracted from published GSK lead-optimization studies
targeting the second bromodomains of the BET-family proteins BRD2, BRD3, BRD4
and BRDT: the GSK046 series (GSK-S1; \cite{gsk046series}), template-hopping
series (GSK-S2; \cite{messenger2021template}), N-methylpyridine-2-carboxamide
series (GSK-S3; \cite{nmethylpyridineseries}) and DNA-encoded-library-derived
GSK040 series (GSK-S4; \cite{taylor2021del}). Structures and measurements
extracted from the articles and supporting information were processed using the
common standardization protocol in Section~\ref{sec:data-processing}.

GSK-S1 uses CLND kinetic solubility, converted from mass concentration to mol/L
using the standardized neutral-parent molecular weight, and reported as
$\log_{10}$(mol/L). This assay conforms to the kinetic-solubility endpoint
represented in the general ADMET training set: solubility measured after
dilution of a DMSO stock into an aqueous buffer.

GSK-S2 uses artificial-membrane apparent permeability, converted from nm/s to
$\log_{10}$(cm/s). The GSK-S2 source endpoint is membrane apparent
permeability, whereas the deployed general model predicts Caco-2 apparent
permeability. The latter was used as a directional permeability surrogate rather
than as an assay-identical reconstruction of the published endpoint.

GSK-S3 and GSK-S4 use $\log D_{7.4}$, retained as the reported dimensionless
$\log_{10}$ distribution coefficient. The general ADMET model was trained on the
corresponding pH-7.4 $\log D_{7.4}$ endpoint and included chromatographic $\log D_{7.4}$ records
meeting the same endpoint definition.

Each task presents one starter and four measured analogues to the agent and
withholds the remaining published structures. Repeated source measurements were
aggregated by the median; missing endpoint values were excluded without
imputation.

\subsection{AstraZeneca campaign}
\label{sec:az-dataset}

The AstraZeneca campaign targets $\log D_{7.4}$ on an undisclosed internal programme.
The compound set, the target, and the therapeutic area are confidential and are
not released; only aggregate distributions and the campaign objective are
reported here. Reference and held-out cohorts were supplied as measured
$\log D_{7.4}$, aqueous solubility and affinity values without structures, and were
processed under the same unit conventions as
Section~\ref{sec:data-processing}. Cohort sizes and ranges are reported in
Table~\ref{tab:endpoint-composition}; the three endpoints are not measured for all endpoints. The campaign used the
general ADMET stack, and excluded cofolding.

\subsection{RetroDMTA projects}
\label{sec:retrodmta}

The four Sanofi datasets represent projects targeting activated coagulation
factor X (FXa), matrix metalloproteinase-8 (MMP-8), peroxisome
proliferator-activated receptor delta (\PPARd{}), and RENIN
\cite{mas2026retrodmta}. Fixed temporal cutoffs produced the following
pre-cutoff and future cohorts: FXa, 2004-07-28 (803 and 212; 1,015 total);
MMP-8, 2003-04-17 (204 and 226; 430 total); \PPARd{}, 2006-05-08 (355 and 143;
498 total); and RENIN, 2007-02-14 (147 and 241; 388 total).

The primary endpoints are \pKi{} for FXa, \pICfifty{} for MMP-8 and RENIN, and
\pECfifty{} for \PPARd{}, where $\mathrm{p}X = -\log_{10}(X$ in mol/L$)$. FXa
additionally contains aqueous thermodynamic solubility at pH 7.4, supplied in
\textmu{}M and converted as $\log S = \log_{10}$(mol/L), and $\log D_{7.4}$ measured at
pH 7.4. Standardization preserves CPD\_ID and temporal membership. Each CPD\_ID
occurs once, so no repeat aggregation is applied; missing endpoint values are
excluded without imputation.

\begin{table}[htbp]
  \caption{Experimental endpoint composition by dataset and cohort. Ranges and
    medians exclude missing values; no missing endpoint is imputed.}
  \label{tab:endpoint-composition}
  \centering
  \small
  \begin{tabular}{@{}lllrrrl@{}}
    \toprule
    \textbf{Project} & \textbf{Endpoint} & \textbf{Set} & \textbf{N} &
    \textbf{Min} & \textbf{Max} & \textbf{Unit} \\
    \midrule
    GSK-S1 & Kinetic solubility & Reference & 5 & $-3.593$ & $-3.481$ & $\log_{10}$(mol/L) \\
    GSK-S1 & Kinetic solubility & Held out & 31 & $-5.104$ & $-2.882$ & $\log_{10}$(mol/L) \\
    GSK-S2 & Permeability & Reference & 5 & $-5.886$ & $-5.229$ & $\log_{10}$(cm/s) \\
    GSK-S2 & Permeability & Held out & 46 & $-6.523$ & $-4.31$ & $\log_{10}$(cm/s) \\
    GSK-S3 & $\log D_{7.4}$ & Reference & 5 & 3.5 & 4.3 & $\log_{10} D$ \\
    GSK-S3 & $\log D_{7.4}$ & Held out & 38 & 2.2 & 4.6 & $\log_{10} D$ \\
    GSK-S4 & $\log D_{7.4}$ & Reference & 5 & 4.8 & 5.7 & $\log_{10} D$ \\
    GSK-S4 & $\log D_{7.4}$ & Held out & 38 & 1.45 & 5.4 & $\log_{10} D$ \\
    \addlinespace
    AZ & $\log D_{7.4}$ & Reference & 89 & 1.4 & 4.717 & $\log_{10} D$ \\
    AZ & $\log D_{7.4}$ & Held out & 262 & 0.1692 & 5.256 & $\log_{10} D$ \\
    AZ & Aqueous solubility & Reference & 88 & $-7.0$ & $-3.0$ & $\log_{10}$(mol/L) \\
    AZ & Aqueous solubility & Held out & 264 & $-7.0$ & $-3.0$ & $\log_{10}$(mol/L) \\
    AZ & Affinity & Reference & 102 & 4.0 & 8.489 & p-scale \\
    AZ & Affinity & Held out & 843 & 4.0 & 8.497 & p-scale \\
    \addlinespace
    FXA & \pKi{} & Reference & 803 & 4.0 & 10.0 & p-scale \\
    FXA & \pKi{} & Held out & 212 & 4.0 & 10.2218 & p-scale \\
    FXA & Aqueous solubility & Reference & 803 & $-7.0$ & 0.3324 & $\log_{10}$(mol/L) \\
    FXA & Aqueous solubility & Held out & 212 & $-7.0$ & 0.4031 & $\log_{10}$(mol/L) \\
    FXA & $\log D_{7.4}$ & Reference & 803 & $-3.989$ & 6.665 & $\log_{10} D$ \\
    FXA & $\log D_{7.4}$ & Held out & 212 & $-3.62$ & 5.77 & $\log_{10} D$ \\
    MMP8 & \pICfifty{} & Reference & 204 & 4.0 & 8.8861 & p-scale \\
    MMP8 & \pICfifty{} & Held out & 226 & 5.0 & 8.8814 & p-scale \\
    \PPARd{} & \pECfifty{} & Reference & 355 & 4.4815 & 9.8765 & p-scale \\
    \PPARd{} & \pECfifty{} & Held out & 143 & 4.699 & 10.0 & p-scale \\
    RENIN & \pICfifty{} & Reference & 147 & 4.0 & 8.3979 & p-scale \\
    RENIN & \pICfifty{} & Held out & 241 & 4.0 & 9.0 & p-scale \\
    \bottomrule
  \end{tabular}
\end{table}

\subsection{Data processing and sample flow}
\label{sec:data-processing}

Structures were parsed and sanitized with RDKit. Salts, cosolvents and
surfactants were removed, and the largest organic fragment was retained,
selected by heavy atom count and then molecular weight. Supplied stereochemistry
was preserved during aromatic reperception and functional-group normalization.
Structures were converted to the RDKit canonical tautomer, assigned their
dominant ionization state at pH 7.4 and represented as canonical isomeric
SMILES.

Retained structures contained only B, Br, C, Cl, F, H, I, N, O, P, S, Se or Si;
had more than eight heavy atoms; contained carbon as more than 20\% of their
heavy atoms; and had neutral-parent molecular weight below 800~Da. Isotopically
labelled structures, unsupported valence or formal-charge states, and molecules
failing RDKit parsing or sanitization were excluded.

All logarithms are base 10. Solubility was converted to mol/L before
transformation to $\log S = \log_{10}$(mol/L), whereas
mass concentrations were divided by the standardized neutral-parent molecular
weight before transformation. Permeability was converted to cm/s before
transformation to $\log P_{\mathrm{app}} = \log_{10}$(cm/s).
Efflux was represented as
$\log_{10}[P_{\mathrm{app}}(\mathrm{B}\rightarrow\mathrm{A}) /
P_{\mathrm{app}}(\mathrm{A}\rightarrow\mathrm{B})]$, and potency as
$\mathrm{p}X = -\log_{10}(X$ in mol/L$)$. Reported logP, and $\log D_{7.4}$
values were already logarithmic and dimensionless.

For duplicate global-ADMET records, measurements were first converted to the
common endpoint unit. For exactly two measurements, their absolute difference
was required to be below twice the assay-level standard deviation; otherwise,
the group was excluded. Accepted pairs retained the measurement in the
prespecified less-favorable direction---for example, lower solubility or
permeability and higher efflux---to limit false-positive labels. For three or
more measurements, groups were retained only when their within-group standard
deviation was below 0.5 log units, and the observed value closest to the median
was selected. Missing endpoints were never imputed.

Publication or temporal cohort membership was assigned before campaign analysis.
Every submitted proposal remained in the molecular-validity denominator. Invalid
proposals were excluded from fingerprints, descriptors, and model analyses but
were recorded explicitly.

\subsection{Information availability and leakage controls}
\label{sec:leakage}

The temporal split prevents explicit future endpoint values from being supplied
to MAGI but cannot exclude exposure through public literature or model
pretraining. Sanofi project-model training and future cohorts had no identical
InChIKey overlap with a non-missing primary endpoint. For GSK, exact
endpoint-labelled overlap with the global ADMET training collection was asserted to be zero.

To contextualize structure-conditioned predictions, one crystallographic
reference complex was selected per target (Table~\ref{tab:constructs}).
Successfully cofolded MAGI proposals were compared with the selected crystal
ligand using residue-level interaction fingerprints.

\begin{table}[htbp]
  \caption{Protein constructs and selected crystallographic
    reference complexes used for structure-conditioned evaluation.}
  \label{tab:constructs}
  \centering
  \small
  \begin{tabular}{@{}lllllll@{}}
    \toprule
    \textbf{Target} & \textbf{UniProt} & \textbf{Construct aa} & \textbf{PDB} &
    \textbf{Chain} & \textbf{Ligand} & \textbf{Sources} \\
    \midrule
    FXA & P00742 & 232 & 9I24 & H & 0GJ &
      \cite{pinto2007apixaban, pinto2001dpc423, lam2003guanidine,
             difenza2007trypsin, uniprot2023, burley2019rcsb} \\
    MMP8 & P22894 & 163 & 1I76 & A & BSI &
      \cite{uniprot2023, burley2019rcsb, grams1995collagenase, bode1995mmp8,
             matter1999mmp8, fischer2019mmp} \\
    \PPARd{} & Q03181 & 272 & 5U3Q & A & 7UJ &
      \cite{uniprot2023, burley2019rcsb, sznaidman2003ppard, oliver2001ppard,
             wu2017ppard} \\
    RENIN & P00797 & 340 & 3K1W & A & BFX &
      \cite{uniprot2023, burley2019rcsb, webb2010renin, maibaum2007aliskiren,
             sph3127renin, rahuel2000renin} \\
    \bottomrule
  \end{tabular}
\end{table}

\begin{figure}[htbp]
  \centering
  \includegraphics[width=\linewidth]{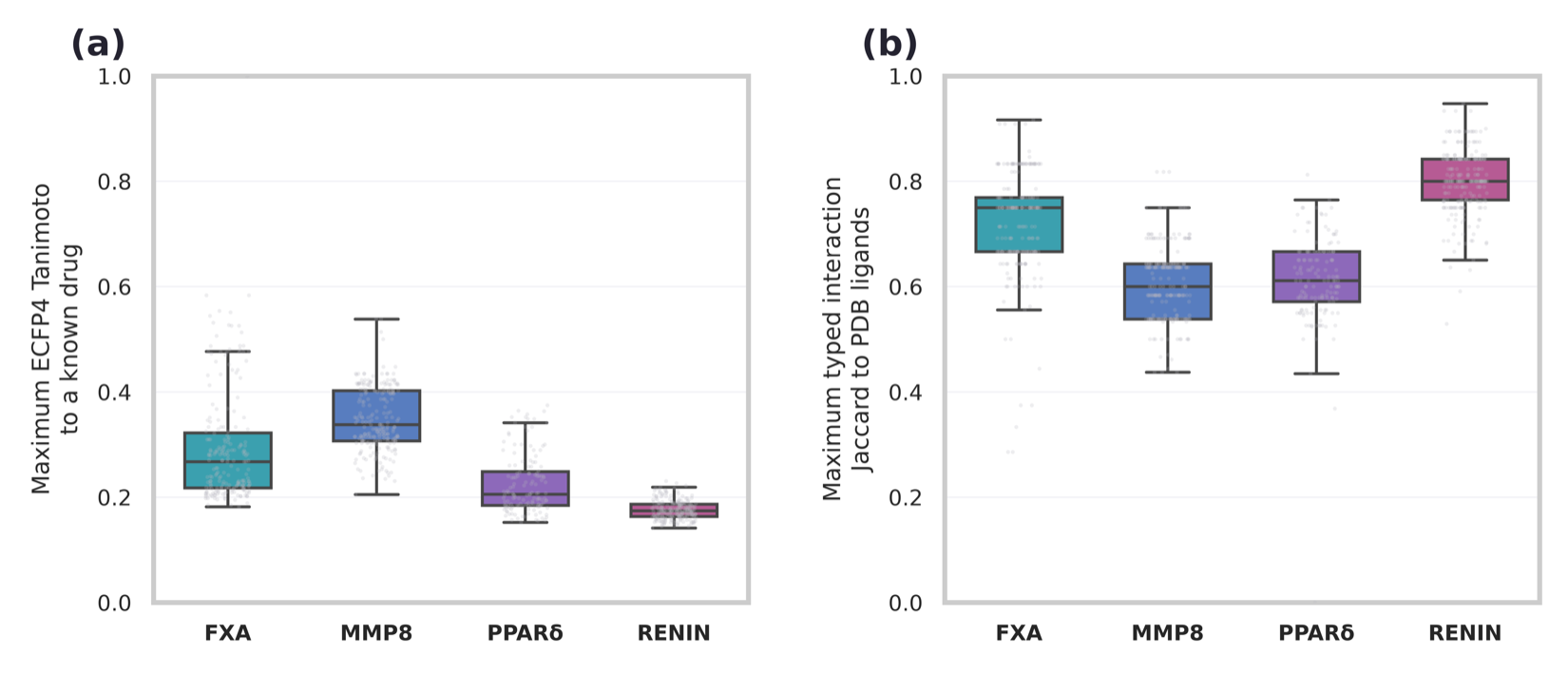}
  \caption{\textbf{Similarity between the held-out set and publicly available
    data (a) ECFP4 similarity to known drugs.} For each FXa, MMP8, \PPARd{}, and
    RENIN held-out compound, points show the maximum ECFP4 Tanimoto similarity
    to a standardized drugs structures extracted from SureChEMBL. \textbf{(b) Interaction-fingerprint similarity to
    crystallographic ligands.} For each FXa, MMP8, \PPARd{}, and RENIN held-out
    compound, points show the maximum interaction fingerprint with Jaccard
    similarity to available experimental structures extracted from the PDB.
    Boxes show medians and interquartile ranges, with whiskers extending to 1.5
    times the interquartile range.}
  \label{fig:heldout-similarity}
\end{figure}

Interaction similarity was the Jaccard overlap between residue-level contact
fingerprints for a generated pose and its target's selected crystallographic
ligand. Buffer components, ions, and incomplete ligands were excluded from the
reference complexes. This analysis measures recovery of the selected binding
interaction pattern \cite{burley2019rcsb}. MMP8 proposals showed the highest
interaction-fingerprint similarity to the selected crystallographic ligand,
followed by \PPARd{} and RENIN, while FXA showed lower overlap
(Figure~\ref{fig:heldout-similarity}b).

\section{Services}
\label{sec:services}

\subsection{Global ADMET models}
\label{sec:global-admet}

The general ADMET ensemble was trained on 156,797 curated molecules carrying
186,120 sparse labels across seven endpoints. XGBoost models use RDKit
physicochemical descriptors concatenated with 2,048-bit Avalon fingerprints
\cite{chen2016xgboost}. Chemprop-MTL uses a depth-3 directed message-passing
network with hidden dimension 300, a two-layer 300-unit FFN, dropout 0.2 and
norm aggregation \cite{yang2019chemprop}. CheMeleon-MTL uses the
foundation-initialized 2,048-dimensional, depth-6 message-passing network with a
task-specific FFN and native aggregation
\cite{kim2023pubchem, burns2026chemeleon}. The multitask networks use MAE in
standardized target space, inverse-standard-deviation task weighting, a
200-epoch cap, and early stopping patience 25. Pooled out-of-fold performance
uses five shared Butina folds; deployed models were retrained on all data. The
consensus is the mean across model families and cross-family standard deviation
reports of disagreement. Performances are reported in
Table~\ref{tab:oof-performance}.

\begin{table}[htbp]
  \caption{Pooled out-of-fold performance on five shared Butina folds. Metrics
    are calculated in each endpoint's native transformed unit; $R^2$ is
    unitless.}
  \label{tab:oof-performance}
  \centering
  \small
  \begin{tabular}{@{}llrrr@{}}
    \toprule
    \textbf{Endpoint} & \textbf{Methods} & \textbf{$R^2$} & \textbf{RMSE} &
    \textbf{MAE} \\
    \midrule
    \multirow{4}{*}{\textbf{LogP}}
      & XGBoost       & 0.771 & 0.855 & 0.605 \\
      & Chemprop-MTL  & 0.803 & 0.793 & 0.540 \\
      & Chemeleon-MTL & 0.792 & 0.815 & 0.578 \\
      & Consensus     & \textbf{0.826} & \textbf{0.744} & \textbf{0.509} \\
    \midrule
    \multirow{4}{*}{\textbf{log}\,$\mathbf{D_{7.4}}$}
      & XGBoost       & 0.618 & 0.860 & 0.639 \\
      & Chemprop-MTL  & 0.720 & 0.736 & 0.525 \\
      & Chemeleon-MTL & 0.704 & 0.758 & 0.544 \\
      & Consensus     & \textbf{0.740} & \textbf{0.709} & \textbf{0.507} \\
    \midrule
    \multirow{4}{*}{\parbox{0.28\linewidth}{\textbf{Aqueous solubility}}}
      & XGBoost       & 0.625 & 0.967 & 0.724 \\
      & Chemprop-MTL  & 0.678 & 0.895 & 0.659 \\
      & Chemeleon-MTL & 0.644 & 0.942 & 0.697 \\
      & Consensus     & \textbf{0.704} & \textbf{0.859} & \textbf{0.630} \\
    \midrule
    \multirow{4}{*}{\textbf{Kinetic solubility}}
      & XGBoost       & 0.301 & 0.512 & 0.396 \\
      & Chemprop-MTL  & 0.417 & 0.468 & 0.325 \\
      & Chemeleon-MTL & 0.378 & 0.483 & 0.349 \\
      & Consensus     & \textbf{0.475} & \textbf{0.444} & \textbf{0.328} \\
    \midrule
    \multirow{4}{*}{\textbf{Permeability Caco-2}}
      & XGBoost       & 0.236 & 0.533 & 0.401 \\
      & Chemprop-MTL  & 0.364 & 0.486 & 0.352 \\
      & Chemeleon-MTL & 0.334 & 0.497 & 0.366 \\
      & Consensus     & \textbf{0.414} & \textbf{0.467} & \textbf{0.342} \\
    \midrule
    \multirow{4}{*}{\textbf{Efflux ratio Caco-2}}
      & XGBoost       & 0.330 & 0.339 & 0.243 \\
      & Chemprop-MTL  & \textbf{0.545} & \textbf{0.280} & \textbf{0.193} \\
      & Chemeleon-MTL & 0.437 & 0.311 & 0.222 \\
      & Consensus     & 0.521 & 0.287 & 0.200 \\
    \midrule
    \multirow{4}{*}{\textbf{Efflux ratio MDCK-MDR1}}
      & XGBoost       & 0.291 & 0.533 & 0.407 \\
      & Chemprop-MTL  & 0.363 & 0.505 & 0.357 \\
      & Chemeleon-MTL & 0.319 & 0.522 & 0.394 \\
      & Consensus     & \textbf{0.438} & \textbf{0.474} & \textbf{0.350} \\
    \bottomrule
  \end{tabular}
\end{table}

\begin{figure}[p]
  \centering
  \includegraphics[width=\linewidth, height=0.84\textheight, keepaspectratio]{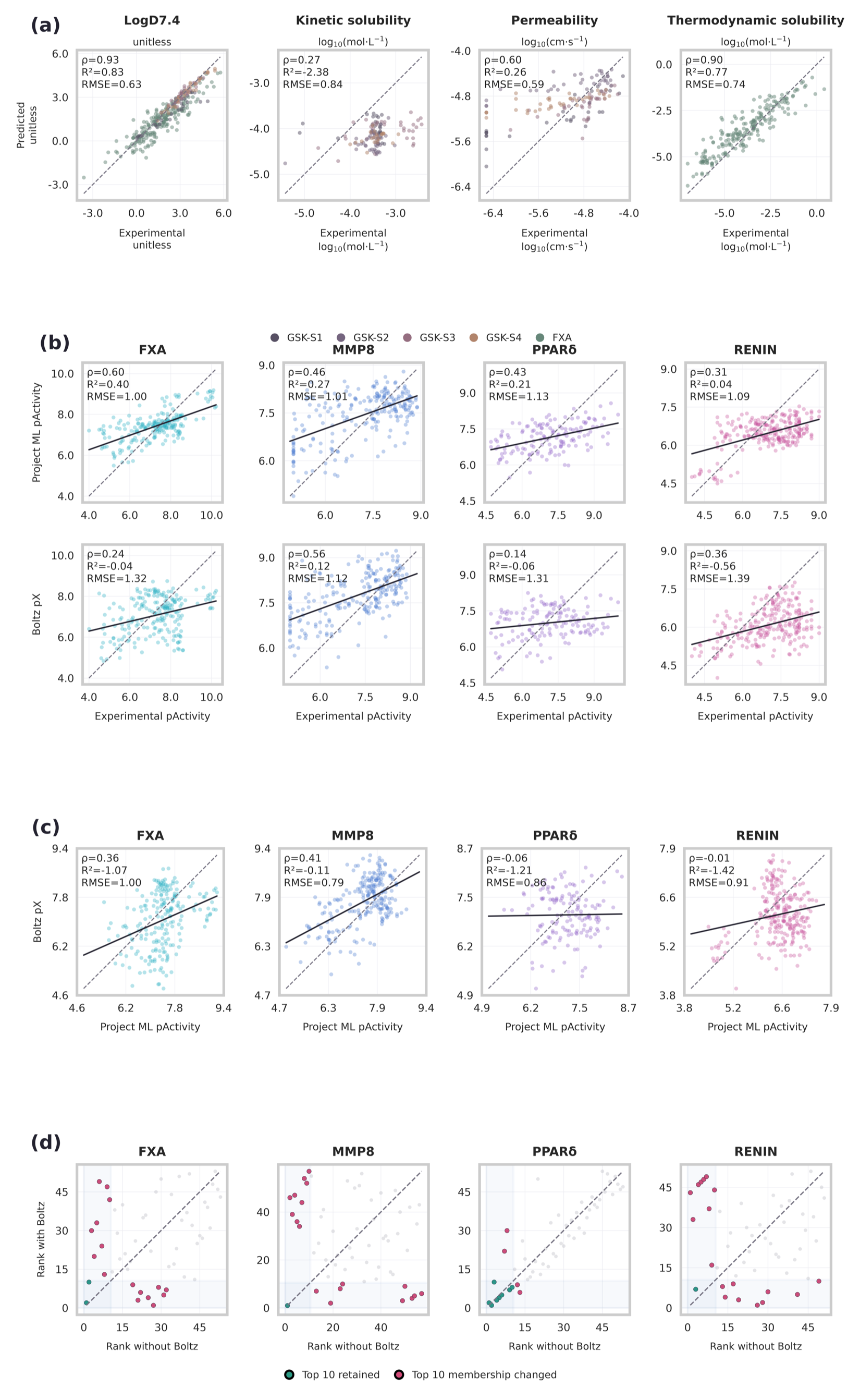}
  \caption{\textbf{Validation of ligand-based models and Boltz-2 on experimental
    data.} (a) Held-out validation of the ADMET models on the projects data. Scatter plots of the prediction from the consensus ADMET model to the GSK and FXA experimental ADMET data. Prediction are grouped by endpoints and colored by projects: GSK-S1 (light blue), GSK-S2 (purple), GSK-S3 (pink), GSK-S4 (orange), and FXA (green). (b) Predictive performance of Sanofi project models (top) and Boltz-2 affinity prediction (bottom) on
    the held-out set: experimental activity versus ligand-based model and Boltz-2
    predictions for FXA, MMP8, \PPARd{} and RENIN. Point are colored based on the target. (c) Agreement between
    ligand-based model and Boltz-2 affinity predictions on held-out Sanofi compounds:
    ligand-based model predictions versus Boltz-2 pX. (d) Post hoc sensitivity of
    proposal ranking to the Boltz-2 affinity term. Compounds were ranked with or without Boltz, evaluating ranking agreement between the two methods. Top 10 ranks with or without Boltz are covered by a grey area. Compounds with matching ranking are colored in green, and otherwise in red.}
    
  \label{fig:model-validation}
\end{figure}

The trained ensemble was applied to a one-million-compound Enamine REAL sample
to calibrate operational similarity thresholds. The fifth-percentile thresholds
are Avalon nearest-neighbour Tanimoto $\geq$0.4358 and Chemprop latent-space
cosine $\geq$0.7334. The deployed overall AD pass requires both criteria;
per-axis scores and pass flags remain available. These thresholds indicate
similarity to model training space and are not calibrated probabilities of
prediction correctness. Held-out endpoint calibration is shown in
Figure~\ref{fig:model-validation}a.

The models are exposed through a persistent REST service that standardizes and
scores each submitted SMILES and returns family predictions, consensus mean,
cross-model disagreement and applicability metadata. The released inference
artifact does not contain row-level endpoint labels, so exact endpoint-specific
GSK training overlap is not claimed.

\subsection{Local project-specific models}
\label{sec:project-models}

\paragraph{Project model recipes.}
Project-specific ligand-based models were developed only for the Sanofi
campaigns as these campaigns included affinity endpoints supported by
sufficiently large quantity of historical measurements for model training. Their primary purpose was to
provide campaign-specific predictions for scoring and prioritization; they also
provided a ligand-based baseline orthogonal to Boltz-2. Per endpoint we trained an ensemble of XGBoost, Chemprop MTL and CheMeleon MTL models; in the multitask objective the
project endpoints carry half of the total task weight and the general ADMET
endpoints enter as auxiliary tasks summing to the other half, up weighting the
scarcer project data (fewer endpoints). Every model sees only compounds before
the task temporal cutoff to prevent leakage and is served exactly like the
global service. The target and temporal window must be fixed explicitly for a
controlled replay; omitting the window causes the service to fall back to its
latest snapshot. The ligand-based models were evaluated on the held-out sets
shown in Figure~\ref{fig:model-validation}b. The ligand-based project models
reach rank correlations above 0.4 for FXA, MMP8 and \PPARd{} (with low $R^2$),
while RENIN shows a low $R^2$ at $\rho \approx 0.31$. Because no error-calibrated
applicability-domain threshold was established for the project-specific models,
cross-model disagreement is treated as a qualitative extrapolation warning and
is not referred to as a formal out-of-domain classification.

\subsection{Boltz-2}
\label{sec:boltz2}

Boltz-2 predicts a protein--ligand complex and an affinity-derived pX
for each submitted structure \cite{passaro2025boltz2}. Sanofi task prompts
designate this affinity as a co-primary objective and requires a
PoseBusters-valid pose as a hard gate. Cofolding uses the fixed full target
sequence and a precomputed MSA, with 3 recycling steps and 200 diffusion
sampling steps (one diffusion sample per structure) followed by affinity
sampling; the service returns the predicted complex, confidence score, ligand
iPTM, affinity value and binder probability. The affinity output,
$\log_{10}(\ICfifty{}/\textrm{\textmu M})$, is converted to the conventional
p-scale. For throughput, the model is kept resident in GPU memory behind a
persistent service, and Boltz-2 runs on a dedicated GPU while REINVENT 4
occupies a separate GPU.

The agreement with the ligand-based model on the held-out set was evaluated in
Figure~\ref{fig:model-validation}c. The two methods agree only for MMP8, whose
project compounds have similar analogues in the PDB---despite different
architectures and non-identical training sets, both encode similar information
for that series. For FXA both models identify strong binders but disagree at the
extremes (most- and least-active), giving poor overall agreement, while for
\PPARd{} and RENIN the predictions do not agree. This pattern reinforces that
ligand- and structure-based predictions are complementary rather than redundant.

\subsection{PoseBusters}
\label{sec:posebusters}

PoseBusters screens the ligand extracted from each Boltz complex in mol
mode, evaluating only ligand geometry and chemistry (bond lengths and angles,
planarity, internal clashes, valences, etc.) without receptor-dependent docking
validation. The score is the fraction of the 12 Boolean checks passed; all\_pass
requires every check and missing or invalid upstream structures to return null
and are not imputed. Observed pass fractions range from 0.833 to 1.0, and a
fraction below 1 is not reported as a complete pass. The same output is used in
two ways: the pass fraction can enter the REINVENT objective as a
structural-plausibility reward, and the per-check breakdown is returned to the
agent as metadata flagging specific geometric/chemical liabilities.

\subsection{Interaction Profiler}
\label{sec:interaction-profiler}

The interaction profiler is an interpretation scoring service. In
structure-plus-SMILES mode it (i) verifies ligand identity against the submitted
SMILES, (ii) prepares and protonates both ligand and receptor at pH 7.4, (iii)
crops an 8~\AA{} binding-site region around the ligand, and (iv) enumerates
protein--ligand contacts within 4~\AA{}, returning typed contacts and a human
readable interaction fingerprint, verifying medicinal chemistry taxonomy of
molecule fragments. Each contact is recorded as a structured triplet linking:
the ligand fragment named and localized under a medicinal-chemistry taxonomy;
the interaction type (e.g., H-bond, hydrophobic, $\pi$-stacking, salt bridge,
halogen bond) together with its distance, angle and geometric validity; the
protein partner residue, the specific interacting atom, and whether it is a
backbone or side chain. This mode can also be used in the objective function as
an additional endpoint by reporting the count of distinct types of interactions.
In SMILES-only mode it returns just the medicinal-chemistry fragment taxonomy.

\section{Metrics}
\label{sec:metrics}

\subsection{Attainment scope and limitations}
\label{sec:attainment-scope}

Primary-endpoint attainment alone does not establish multiparameter success. We
therefore report primary-objective attainment, secondary-objective attainment,
fixed-guardrail compliance, and complete TTP attainment separately. A molecule
attains the complete TTP only when all required objectives and guardrails pass.

For global ADMET models, predictions are classified as in domain, singly
flagged, or outside the operational applicability domain according to the two
fixed similarity rules defined in Section~\ref{sec:global-admet}. Numerical
predictions are retained for transparency but are flagged in analyses;
attainment is summarized both overall and, where sample size permits, among
in-domain predictions only. For Sanofi project-specific models, no formal
 applicability-domain classifier was implemented. Cross-model standard
deviation is therefore reported as an extrapolation warning rather than used to
classify a prediction as formally in or out of domain.

\subsection{Molecular quality}
\label{sec:molecular-quality}

Novelty is 1 minus the maximum Tanimoto similarity over ECFP4 fingerprints
\cite{rogers2010ecfp} to the shown set; internal diversity is 1 minus the mean
pairwise Tanimoto; synthesizability is the synthetic-accessibility score
\cite{ertl2009sascore}; scaffold provenance uses Bemis--Murcko frameworks
\cite{bemis1996scaffolds}; all cheminformatics uses RDKit \cite{rdkit}. Calibration is
predicted against measured agreement on held-out compounds, reported as Pearson
$r$, mean absolute error and bias at the withheld optimized compound. The
cofolding ablation re-ranks the same cofolded molecules by the ligand-only
objective and by the objective including the Boltz-2 term, comparing orderings
by Spearman rank correlation and top K overlap.

\subsection{GSK and AstraZeneca primary-endpoint outcomes}
\label{sec:gsk-primary-endpoints}

\begin{figure}[p]
  \centering
  \includegraphics[width=\linewidth, height=0.9\textheight, keepaspectratio]{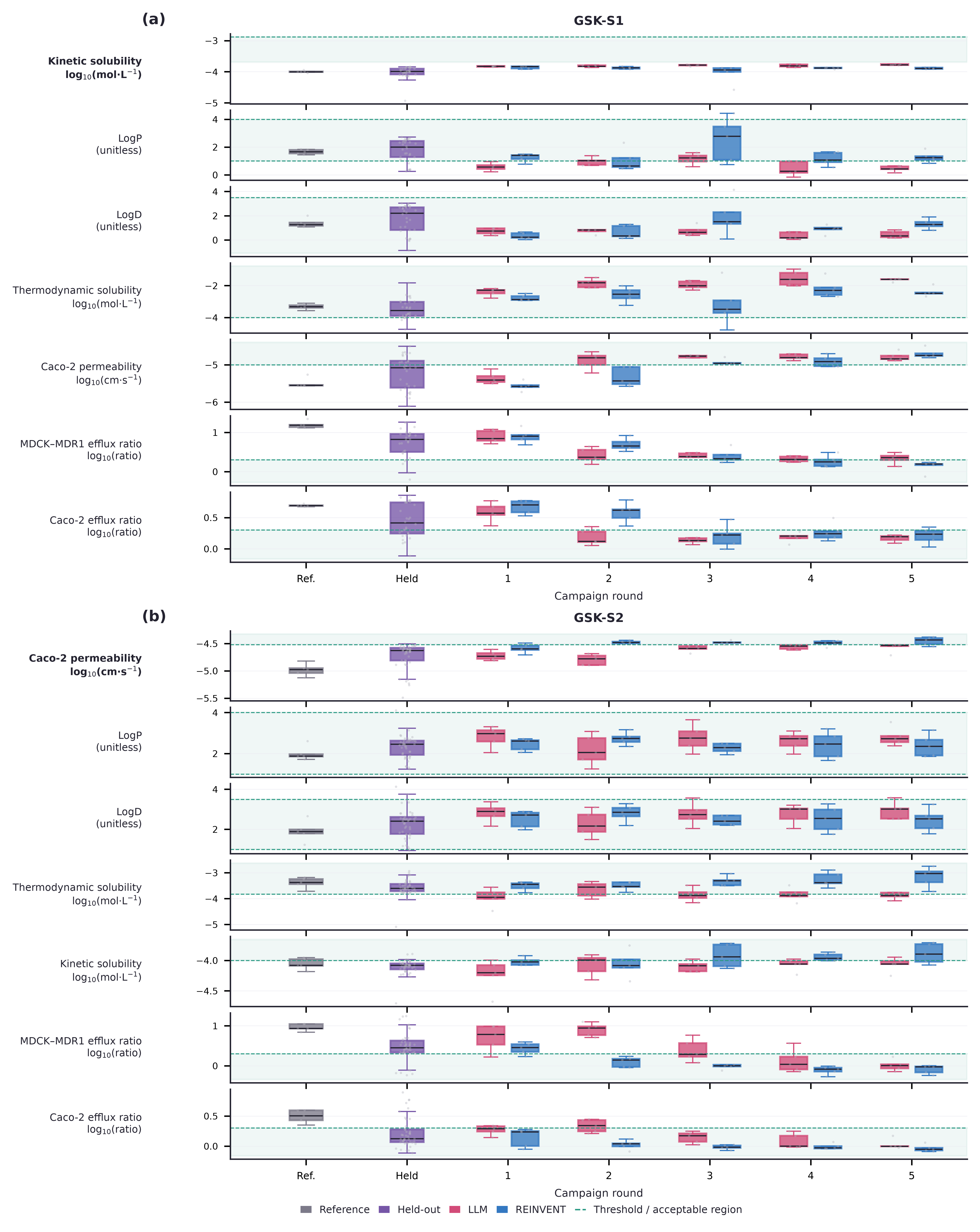}
  \caption{\textbf{Property distributions during rounds 1--5 of the GSK-S1 and
    GSK-S2 campaigns.} Gray denotes the five shown Reference compounds, green
    for the published Held-out compounds, red LLM proposals and blue final
    agent-selected REINVENT proposals. Points represent individual molecules;
    boxes show the median and interquartile range, with whiskers extending 1.5
    times the interquartile range. Each row identifies the endpoint and unit.
    Dashed green lines mark project-specific criteria where available and
    generic ADMET context ranges otherwise; pale-green shading shows the
    corresponding favorable region. Reference and Held-out values are study
    measurements, whereas proposal values are model predictions.}
  \label{fig:gsk12-properties}
\end{figure}

Figure~\ref{fig:main-quality}b reports ligand-model primary-endpoint
distributions for the four GSK projects and the AstraZeneca campaign. Three of
four GSK project produced at least one proposal satisfying
the model-space primary threshold: REINVENT for GSK-S2 and both routes for
GSK-S3 and GSK-S4. Neither route met the GSK-S1 kinetic-solubility objective,
and LLM remained below the GSK-S2 permeability criterion. GSK-S1 held-out
agreement was weak and range-compressed
(Table~\ref{tab:oof-performance}; Figure~\ref{fig:model-validation}a), making
predictions near its target extrapolative. The agent consequently treated
aqueous solubility as a secondary objective, where improvement was clearer
(Figure~\ref{fig:gsk12-properties}).

\subsection{Statistical analysis}
\label{sec:statistics}

\begin{figure}[p]
  \centering
  \includegraphics[width=\linewidth, height=0.9\textheight, keepaspectratio]{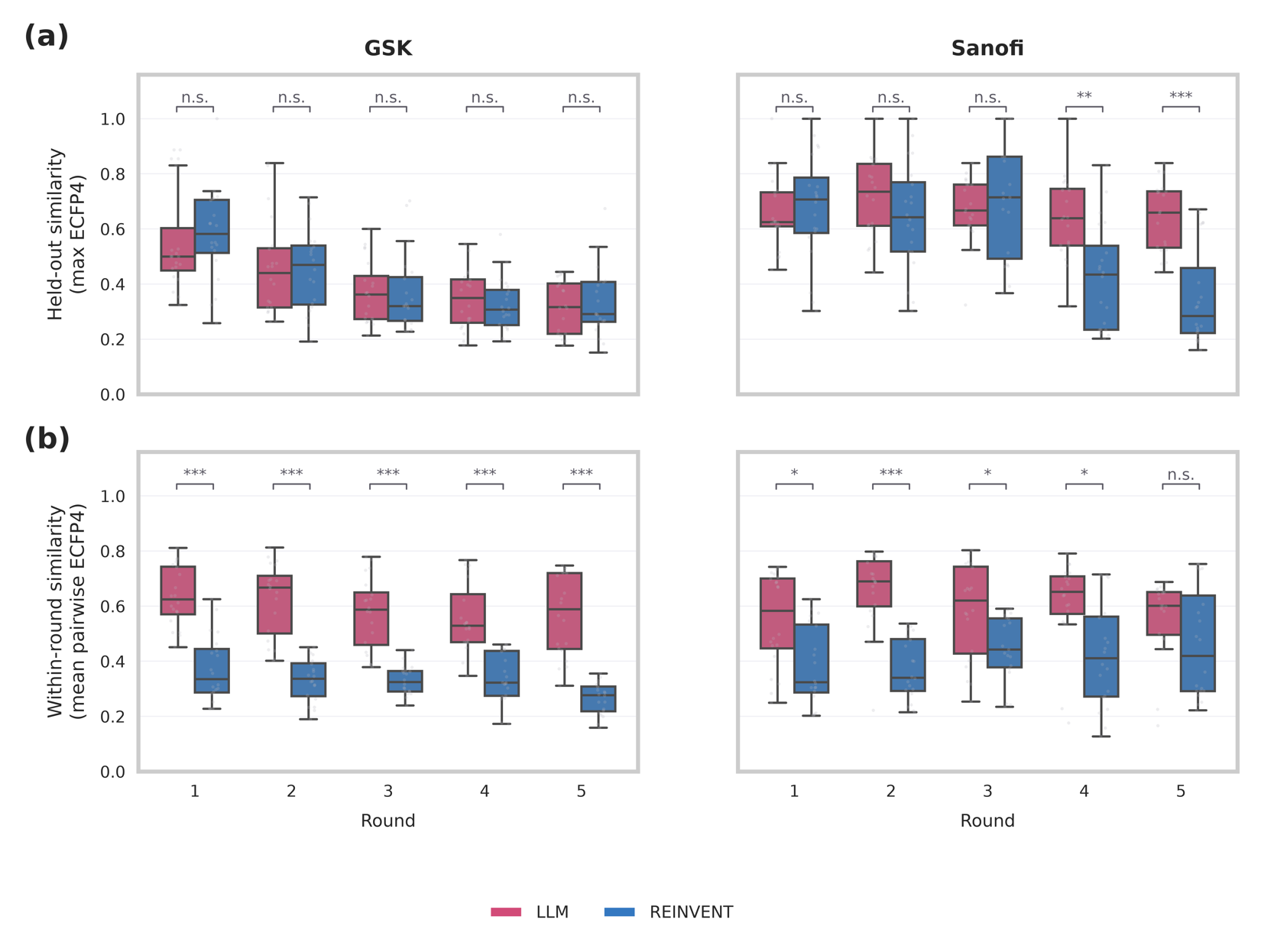}
  \caption{\textbf{Round-by-round similarity to held-out chemistry and
    intra-round similarity.} Red boxes represent LLM proposals, and blue boxes
    represent final agent-selected REINVENT proposals during rounds 1--5.
    Columns separate GSK and Sanofi projects. For each molecule, the upper row
    reports its maximum ECFP4 Tanimoto similarity to any project-specific
    held-out compound. The lower row reports its mean pairwise ECFP4 Tanimoto
    similarity to other molecules from the same project, route, and round; lower
    values therefore indicate greater intra-round diversity. Boxes show medians
    and interquartile ranges; individual outliers are not shown. Brackets
    compare LLM and REINVENT molecule-level distributions within each round
    using two-sided Mann--Whitney U tests, with Holm correction across the five
    rounds separately for each organization and metric. n.s., adjusted
    $p \geq 0.05$; $*$ for $p < 0.05$; $**$ for $p < 0.01$; $***$ for
    $p < 0.001$.}
  \label{fig:round-similarity}
\end{figure}

\begin{figure}[p]
  \centering
  \includegraphics[width=\linewidth, height=0.9\textheight, keepaspectratio]{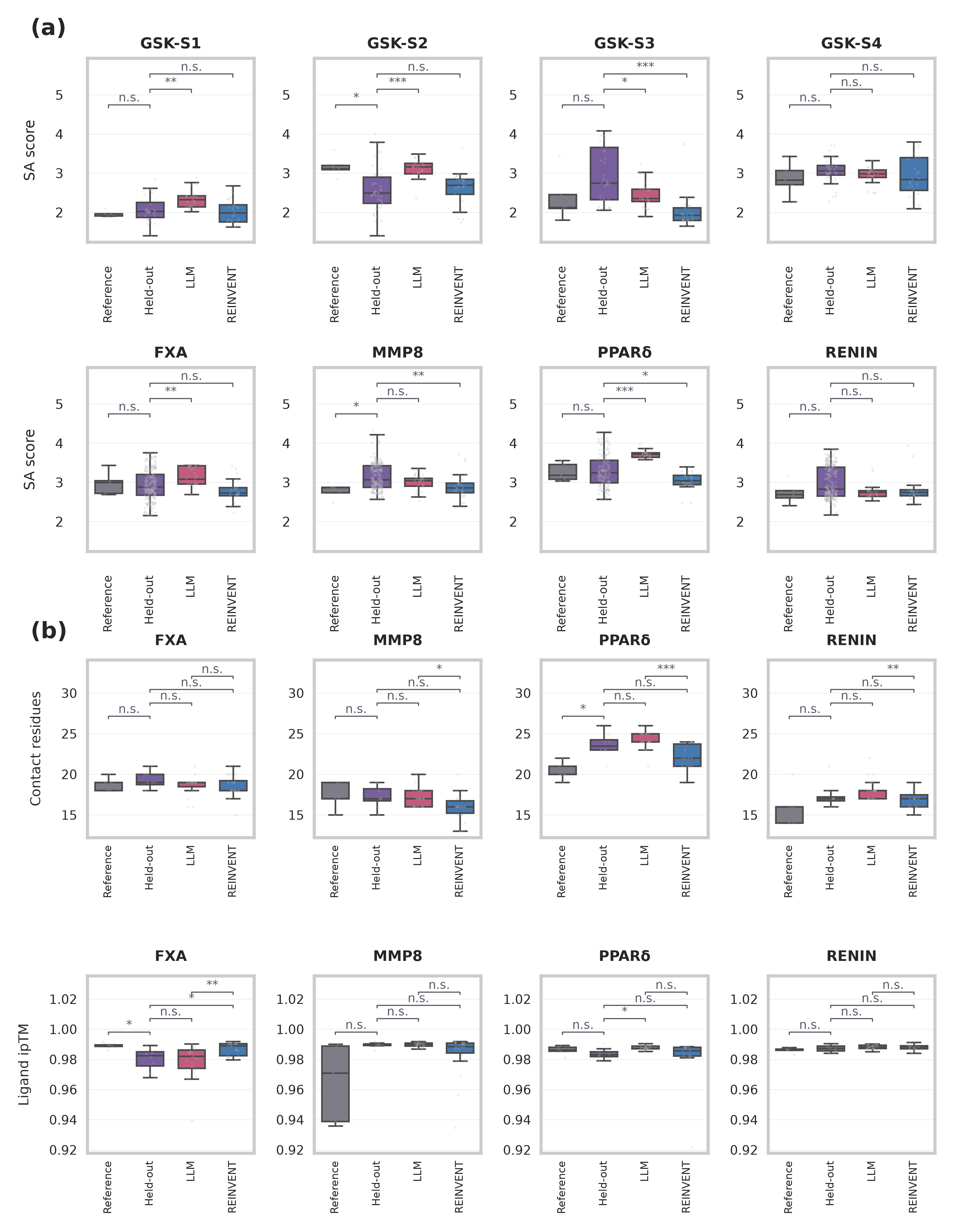}
  \caption{\textbf{Structural quality of the reference held out and proposed
    compounds. (a) Synthetic accessibility (SA) scores across the eight
    projects.} Gray, green, red, and blue denote Reference, Held-out, LLM and
    REINVENT molecules, respectively. (b) \textbf{Structural quality of
    predicted complexes.} Rows report the number of protein residues with a
    heavy atom within 4~\AA{} of the ligand and ligand ipTM. Boxes show medians
    and interquartile ranges; whiskers extend to 1.5 times the interquartile
    ranges. Brackets compare Held-out--LLM, Held-out--REINVENT and
    LLM--REINVENT distributions using two-sided Mann--Whitney U tests. n.s.,
    adjusted $p \geq 0.05$; $*$ for $p < 0.05$; $**$ for $p < 0.01$; $***$ for
    $p < 0.001$.}
  \label{fig:structural-quality}
\end{figure}

For Figure~\ref{fig:main-quality}a, each metric is averaged across
rounds 1--5 within each project and route; LLM and REINVENT project means are
compared with a two-sided Wilcoxon signed-rank test (four paired projects per
organization), with Holm adjustment across the predefined tests. Project-round
points remain visible as descriptive data. Within-project molecule-level
Mann--Whitney comparisons in Figure~\ref{fig:main-quality}b--c and
Figures~\ref{fig:round-similarity} and~\ref{fig:structural-quality} are
exploratory: canonical structures are deduplicated, but analogues remain
chemically dependent, so their p-values are not interpreted as independent
cross-project evidence. Chemist-evaluation proportions
(Figure~\ref{fig:medchem-evaluation}) are reported with Wilson 95\% intervals.
\section{Validation protocol}
\label{sec:validation}

\subsection{Project objectives}
\label{sec:project-objectives}

Each task provides a starter with measured ADMET, four measured reference
analogues, an endpoint objective as a delta on the starter, and guardrails;
potency is out of scope. S1: solubility from 330~\textmu{}M ($\log -3.48$),
raise by at least 0.6 log and permeability to at least 100~nm/s, with
CHI/$\log D_{7.4}$ below 3.5 and MW below 440 \cite{gsk046series}. S2: apparent
permeability from 17~nm/s ($\log -5.77$), raise by at least 1.25 log, with
solubility at least 150~\textmu{}M, no added hydrogen-bond donors and MW below
470 \cite{messenger2021template}. S3: $\log D_{7.4}$ from 4.3, lower by at least
1.1 log, with solubility at least 1000~\textmu{}M, permeability at least
150~nm/s and MW below 420 \cite{nmethylpyridineseries}. S4: $\log D_{7.4}$ from 5.7,
lower by at least 2.85 log, with solubility at least 265~\textmu{}M,
permeability at least 100~nm/s and MW below 470 \cite{taylor2021del}. AZ: $\log D_{7.4}$ lowered below 3.0, with rotatable bonds at most 8. The agent
receives the starter and analogue measurements, and never saw the held-out compounds.

\clearpage

{
\footnotesize
\setlength{\tabcolsep}{3pt}
\begin{longtable}{@{}L{0.08\linewidth}L{0.12\linewidth}L{0.20\linewidth}L{0.09\linewidth}R{0.11\linewidth}R{0.12\linewidth}L{0.13\linewidth}@{}}
  \caption{Project objectives, guardrails, scoring directions, target bounds,
    and required evaluations. Upper desirability values for increasing potency
    objectives are saturation points rather than exclusion thresholds.}
  \label{tab:objectives} \\
  \toprule
  \textbf{Project} & \textbf{Requirement} & \textbf{Endpoint or constraint} &
  \textbf{Direction} & \textbf{Minimum acceptable value} &
  \textbf{Desirability saturation value} & \textbf{Unit} \\
  \midrule
  \endfirsthead
  \toprule
  \textbf{Project} & \textbf{Requirement} & \textbf{Endpoint or constraint} &
  \textbf{Direction} & \textbf{Minimum acceptable value} &
  \textbf{Desirability saturation value} & \textbf{Unit} \\
  \midrule
  \endhead
  \bottomrule
  \endfoot
  GSK-S1 & Primary objective & Kinetic solubility & Increase & $-2.881$ & -- & $\log_{10}$(mol/L) \\
  GSK-S1 & Fixed guardrail & Permeability & Minimum & $-5.0$ & -- & $\log_{10}$(cm/s) \\
  GSK-S1 & Fixed guardrail & $\log D_{7.4}$ & Maximum & -- & 3.5 & $\mathrm{Log}_{10}(D)$ \\
  GSK-S1 & Fixed guardrail & Molecular weight & Maximum & -- & 440.0 & g/mol \\
  \addlinespace
  GSK-S2 & Primary objective & Permeability & Increase & $-4.52$ & -- & $\log_{10}$(cm/s) \\
  GSK-S2 & Fixed guardrail & Aqueous solubility & Minimum & 150.0 & -- & \textmu{}M \\
  GSK-S2 & Fixed guardrail & Hydrogen-bond donors & No increase & -- & -- & count \\
  GSK-S2 & Fixed guardrail & Molecular weight & Maximum & -- & 470.0 & g/mol \\
  \addlinespace
  GSK-S3 & Primary objective & $\log D_{7.4}$ & Decrease & -- & 3.2 & $\mathrm{Log}_{10}(D)$ \\
  GSK-S3 & Fixed guardrail & Aqueous solubility & Minimum & 1000.0 & -- & \textmu{}M \\
  GSK-S3 & Fixed guardrail & Permeability & Minimum & 150.0 & -- & nm/s \\
  GSK-S3 & Fixed guardrail & Molecular weight & Maximum & -- & 420.0 & g/mol \\
  \addlinespace
  GSK-S4 & Primary objective & $\log D_{7.4}$ & Decrease & -- & 2.85 & $\mathrm{Log}_{10}(D)$ \\
  GSK-S4 & Fixed guardrail & Aqueous solubility & Minimum & 265.0 & -- & \textmu{}M \\
  GSK-S4 & Fixed guardrail & Permeability & Minimum & 100.0 & -- & nm/s \\
  GSK-S4 & Fixed guardrail & Molecular weight & Maximum & -- & 470.0 & g/mol \\
  \addlinespace
  AZ & Primary objective & $\log D_{7.4}$ & Decrease & -- & 3.0 & $\mathrm{Log}_{10}(D)$ \\
  AZ & Fixed guardrail & Rotatable bonds & Maximum & -- & 8.0 & count \\
  AZ & Fixed guardrail & Molecular weight & In range & 250.0 & 475.0 & g/mol \\
  \addlinespace
  FXA & Primary objective & \pKi{} & Increase & 8.5 & 9.0 & p-scale \\
  FXA & Secondary objective & Aqueous solubility & Increase & $-4.301$ & $-4.0$ & $\log_{10}$(mol/L) \\
  FXA & Secondary objective & $\log D_{7.4}$ & In range & 1.5 & 3.5 & $\mathrm{Log}_{10}(D)$ \\
  FXA & Fixed guardrail & Molecular weight & In range & 400.0 & 675.0 & g/mol \\
  FXA & Required evaluation & Boltz-2 affinity and PoseBusters validity & Required & -- & -- & p-scale \\
  \addlinespace
  MMP8 & Primary objective & \pICfifty{} & Increase & 8.5 & 9.0 & p-scale \\
  MMP8 & Fixed guardrail & Molecular weight & In range & 350.0 & 575.0 & g/mol \\
  MMP8 & Required evaluation & Boltz-2 affinity and PoseBusters validity & Required & -- & -- & p-scale \\
  \addlinespace
  \PPARd{} & Primary objective & \pECfifty{} & Increase & 8.0 & 9.0 & p-scale \\
  \PPARd{} & Fixed guardrail & Molecular weight & In range & 425.0 & 625.0 & g/mol \\
  \PPARd{} & Required evaluation & Boltz-2 affinity and PoseBusters validity & Required & -- & -- & p-scale \\
  \addlinespace
  RENIN & Primary objective & \pICfifty{} & Increase & 8.0 & 9.0 & p-scale \\
  RENIN & Fixed guardrail & Molecular weight & In range & 325.0 & 525.0 & g/mol \\
  RENIN & Required evaluation & Boltz-2 affinity and PoseBusters validity & Required & -- & -- & p-scale \\
\end{longtable}
}

\subsection{Agent experiment}
\label{sec:agent-experiment}

The LLM route asks Claude Sonnet 4.6 to return approximately five final
structures per round. The REINVENT route creates or updates a TTP, invokes
REINVENT 4, monitors or recovers the run, retrieves its pool, and selects five
final structures. All reported route figures and statistics use the common
complete rounds 1--5 window. GSK and AstraZeneca experiments used the same ADMET evaluators and
exclude cofolding. Sanofi prompts additionally designate project-model
prediction and Boltz-2 affinity as co-primary, a PoseBusters-valid pose as a
hard gate, and interaction analysis as informative metadata.

\begin{table}[htbp]
  \caption{Project-level REINVENT generation pool, shortlist and cofold
    coverage.}
  \label{tab:pool}
  \centering
  \footnotesize
  \setlength{\tabcolsep}{3pt}
  \begin{tabular}{@{}L{0.09\linewidth}R{0.10\linewidth}R{0.11\linewidth}R{0.10\linewidth}L{0.22\linewidth}R{0.11\linewidth}R{0.10\linewidth}@{}}
    \toprule
    \textbf{Project} & \textbf{Pool records n} &
    \textbf{Valid distinct structures} & \textbf{Unique shortlisted} &
    \textbf{Shortlist construction} & \textbf{Cofolded structures} &
    \textbf{Cofold fraction} \\
    \midrule
    GSK-S1 & 96061 & 94363 & 25 & Agent-authored top five list & -- & -- \\
    GSK-S2 & 132765 & 131030 & 25 & Agent-authored top five list & -- & -- \\
    GSK-S3 & 200977 & 195351 & 20 & Agent-authored top five list & -- & -- \\
    GSK-S4 & 204769 & 198243 & 25 & Agent-authored top five list & -- & -- \\
    AZ & 163840 &  -- & 25 & Agent-authored top five list & -- & -- \\
    \addlinespace
    FXA & 63455 & 63455 & 21 & Agent-authored final top five from transcript & 61260 & 0.965 \\
    MMP8 & 117061 & 117061 & 22 & Agent-authored final top five from transcript & 114250 & 0.976 \\
    \PPARd{} & 29328 & 29328 & 18 & Agent-authored final top five from transcript & 28345 & 0.966 \\
    RENIN & 74489 & 74489 & 25 & Agent-authored final top five from transcript & 68266 & 0.916 \\
    \bottomrule
  \end{tabular}
\end{table}

\clearpage

{
\footnotesize
\setlength{\tabcolsep}{3pt}
\begin{longtable}{@{}L{0.10\linewidth}R{0.06\linewidth}R{0.13\linewidth}R{0.15\linewidth}R{0.14\linewidth}R{0.14\linewidth}R{0.10\linewidth}@{}}
  \caption{Exact per-invocation of REINVENT parameters parsed from audited
    operation records.}
  \label{tab:reinvent-params} \\
  \toprule
  \textbf{Project} & \textbf{Round} & \textbf{Invocation ID in round} &
  \textbf{Maximum steps} & \textbf{Batch size} & \textbf{Sigma} &
  \textbf{Seed structures n} \\
  \midrule
  \endfirsthead
  \toprule
  \textbf{Project} & \textbf{Round} & \textbf{Invocation ID in round} &
  \textbf{Maximum steps} & \textbf{Batch size} & \textbf{Sigma} &
  \textbf{Seed structures n} \\
  \midrule
  \endhead
  \bottomrule
  \endfoot
  \textbf{GSK-S1} & 1 & 1 & 75 & 64 & 120 & 5 \\
  \textbf{GSK-S1} & 2 & 1 & 80 & 64 & 130 & 5 \\
  \textbf{GSK-S1} & 3 & 1 & 80 & 64 & 130 & 5 \\
  \textbf{GSK-S1} & 4 & 1 & 80 & 64 & 120 & 5 \\
  \textbf{GSK-S1} & 5 & 1 & 80 & 64 & 115 & 6 \\
  \addlinespace
  \textbf{GSK-S3} & 1 & 1 & 75 & 96 & 120 & 5 \\
  \textbf{GSK-S3} & 2 & 1 & 85 & 96 & 140 & 5 \\
  \textbf{GSK-S3} & 3 & 1 & 90 & 96 & 130 & 5 \\
  \textbf{GSK-S3} & 4 & 1 & 80 & 96 & 140 & 5 \\
  \textbf{GSK-S3} & 5 & 1 & 75 & 96 & 140 & 5 \\
  \addlinespace
  \textbf{GSK-S2} & 1 & 1 & 75 & 64 & 120 & 5 \\
  \textbf{GSK-S2} & 2 & 1 & 80 & 64 & 128 & 5 \\
  \textbf{GSK-S2} & 3 & 1 & 90 & 64 & 140 & 5 \\
  \textbf{GSK-S2} & 4 & 1 & 90 & 64 & 150 & 5 \\
  \textbf{GSK-S2} & 5 & 1 & 90 & 64 & 160 & 5 \\
  \addlinespace
  \textbf{GSK-S4} & 1 & 1/2 & 75/75 & 64/64 & 120/120 & 5 \\
  \textbf{GSK-S4} & 2 & 1 & 80 & 64 & 130 & 5 \\
  \textbf{GSK-S4} & 3 & 1/2 & 80/80 & 64/64 & 130/130 & 5 \\
  \textbf{GSK-S4} & 4 & 1 & 80 & 64 & 130 & 5 \\
  \textbf{GSK-S4} & 5 & 1 & 80 & 64 & 130 & 5 \\
  \addlinespace
  \textbf{AZ} & 1 & 1 & 80 & 384 & 128 & 5 \\
  \textbf{AZ} & 2 & 1 & 80 & 704 & 128 & 5 \\
  \textbf{AZ} & 3 & 1 & 100 & 384 & 128 & 5 \\
  \textbf{AZ} & 4 & 1 & 100 & 256 & 128 & 5 \\
  \textbf{AZ} & 5 & 1 & 100 & 128 & 128 & 5 \\
  \addlinespace
  \textbf{FXA} & 1 & 1/2 & 300/300 & 64/64 & 120/128 & 5 \\
  \textbf{FXA} & 2 & 1 & 300 & 320 & 120 & 2 \\
  \textbf{FXA} & 3 & 1/2 & 300/300 & 64/64 & 128/128 & 2 \\
  \textbf{FXA} & 4 & 1 & 300 & 64 & 120 & 5 \\
  \textbf{FXA} & 5 & 1 & 300 & 64 & 120 & 5 \\
  \addlinespace
  \textbf{MMP8} & 1 & 1/2/3 & 200/200/200 & 32/32/32 & 120/120/120 & 5 \\
  \textbf{MMP8} & 2 & 1/2 & 300/300 & 128/128 & 128/128 & 5 \\
  \textbf{MMP8} & 3 & 1 & 250 & 64 & 120 & 5 \\
  \textbf{MMP8} & 4 & 1 & 150 & 64 & 120 & 6 \\
  \textbf{MMP8} & 5 & 1 & 200 & 128 & 128 & 5 \\
  \addlinespace
  \textbf{\PPARd{}} & 1 & 1 & 200 & 64 & 120 & 5 \\
  \textbf{\PPARd{}} & 2 & 1/2 & 250/250 & 64/32 & 120/120 & 5 \\
  \textbf{\PPARd{}} & 2 & 2 & 250 & 32 & 120 & 5 \\
  \textbf{\PPARd{}} & 3 & 1 & 300 & 64 & 120 & 5 \\
  \textbf{\PPARd{}} & 4 & 1/2/3 & 300/300/300 & 64/64/32 & 120/120/120 & 6 \\
  \textbf{\PPARd{}} & 5 & 1 & 300 & 80 & 120 & 2 \\
  \addlinespace
  \textbf{RENIN} & 1 & 1/2/3 & 200/200/200 & 32/32/32 & 120/120/120 & 5 \\
  \textbf{RENIN} & 2 & 1/2/3 & 200/200/200 & 128/128/128 & 120/120/120 & 5 \\
  \textbf{RENIN} & 3 & 1 & 200 & 64 & 120 & 3 \\
  \textbf{RENIN} & 4 & 1/2 & 250/250 & 128/128 & 110/110 & 5 \\
  \textbf{RENIN} & 5 & 1/2 & 300/300 & 96/96 & 120/120 & 5 \\
\end{longtable}
}

\subsection{Medicinal chemist evaluation}
\label{sec:medchem-eval}

The prepared blinded instrument contains 28 questions: 12 source-discrimination
cases, 11 structure-first candidate-prioritization cases, and 5 SAR-reasoning
cases. Source-discrimination cases ask which of one agent's proposal, and one
similarity-matched held-out compound is AI-generated. Prioritization cases ask
the evaluator to select among blinded LLM, REINVENT, and held-out candidates
under an explicit property objective. SAR cases request a verdict of supported,
plausible but incomplete, or unsupported, together with confidence and an
optional negative-reason code.

\begin{figure}[htbp]
  \centering
  \includegraphics[width=\linewidth]{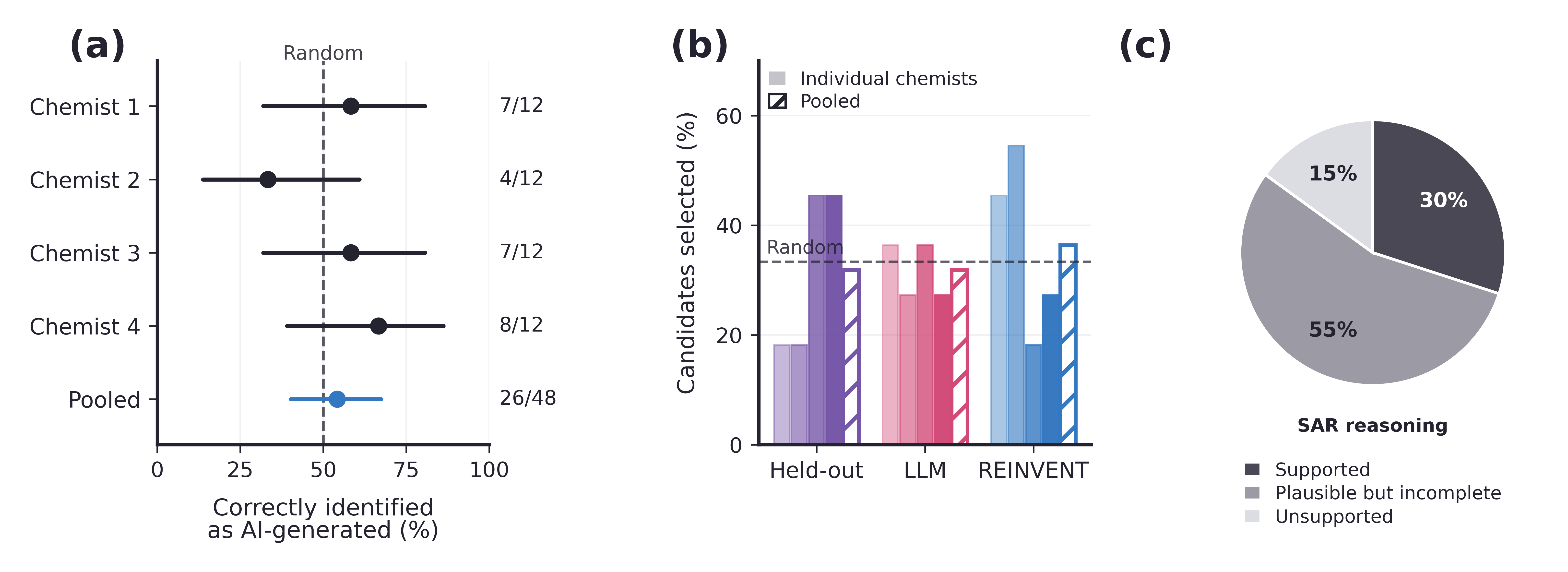}
  \caption{\textbf{Blinded medicinal-chemist evaluation of agent proposals.}
    Four chemists completed the 28-question instrument described in
    Section~\ref{sec:medchem-eval}. \textbf{(a) Source discrimination.} For each
    of the 12 paired cases, the evaluator judged which of two structures was
    AI-generated; points show the proportion correct with Wilson 95\% intervals,
    and the dashed line marks chance. Counts to the right give correct answers
    over cases answered. \textbf{(b) Candidate prioritization.} Proportion of the
    11 prioritization cases in which a held-out, LLM or REINVENT candidate was
    selected as the next compound to test; solid bars are individual chemists
    and the hatched bar pools all responses. The dashed line marks the
    proportion expected if the three sources were chosen at random. \textbf{(c) SAR reasoning.} Distribution of verdicts over the 5 rationales, pooled across chemists.}
  \label{fig:medchem-evaluation}
\end{figure}

Primary analyses are discrimination accuracy with a binomial 95\% interval
relative to chance, candidate-choice proportions, and reasoning-verdict
proportions. Results for the four chemists who
completed the instrument are shown in
Figure~\ref{fig:medchem-evaluation}. Discrimination accuracy did not separate
from chance for any chemist or in pooled analysis, and no source was
preferentially selected in the prioritization cases.

\section{Availability and ethics}
\label{sec:availability}

\subsection{Human-participant ethics}
\label{sec:ethics}

The blinded medicinal-chemist evaluation involved 4 participants. No personally identifying participant
information is released.

\subsection{Responsible release}
\label{sec:responsible-release}

Molecular design systems have potential dual-use applications and may generate
unsafe, reactive, or otherwise inappropriate structures. MAGI outputs are
predictions rather than experimentally validated compounds and require expert
review.

\section{Supplementary figures}
\label{sec:supplementary-figures}

Extended figures supporting the main text analyses are shown below. Main-text
cross-references identify the relevant panels; underlying numerical values and
tests are provided in the accompanying data package.

\begin{figure}[p]
  \centering
  \includegraphics[width=\linewidth, height=0.9\textheight, keepaspectratio]{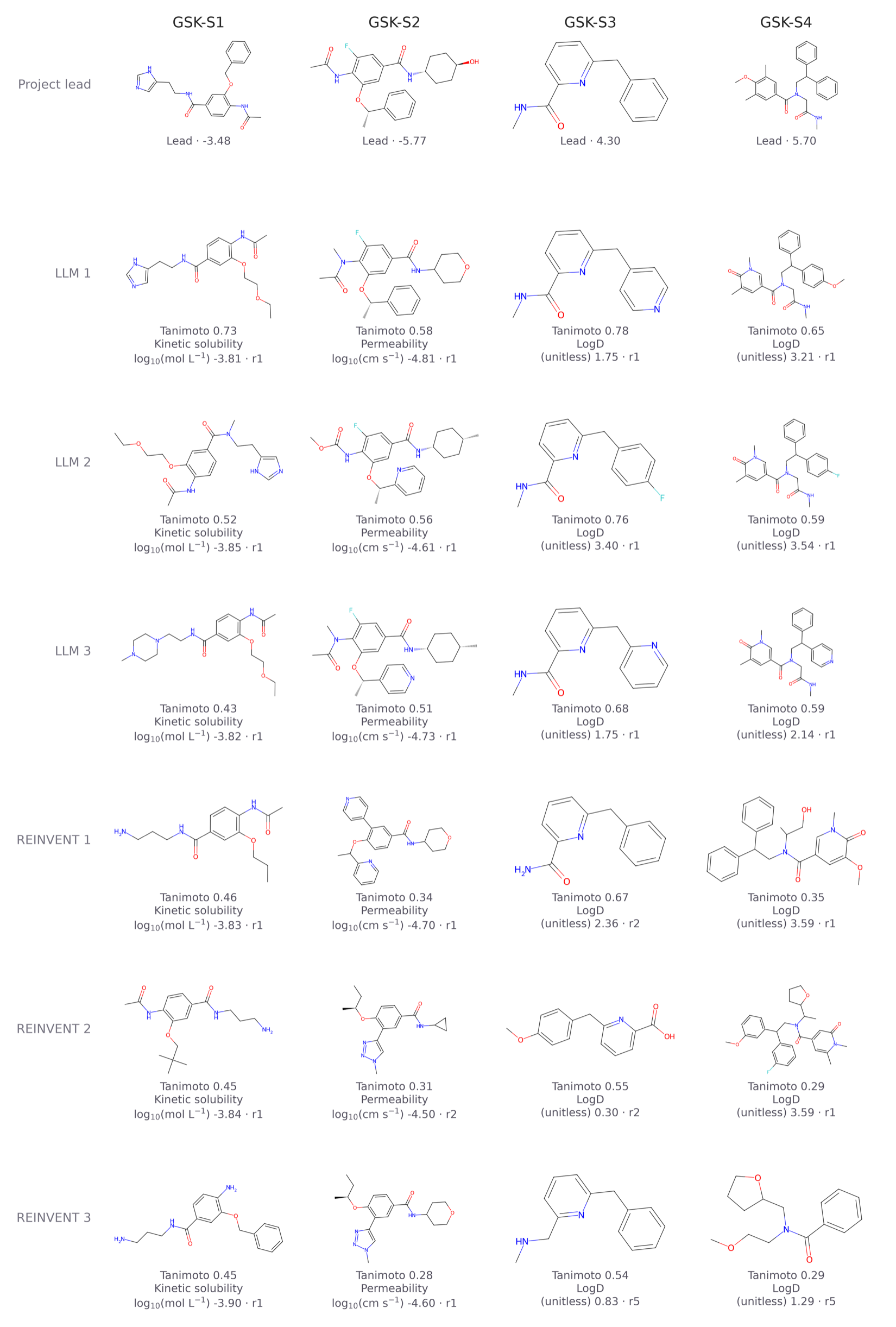}
  \caption{\textbf{Best held-out compounds and top-ranked LLM and REINVENT
    proposals for the GSK projects.} Best measured held-out molecule plus the
    top three LLM and top three REINVENT proposals per project.}
  \label{fig:gsk-best}
\end{figure}

\begin{figure}[p]
  \centering
  \includegraphics[width=\linewidth, height=0.9\textheight, keepaspectratio]{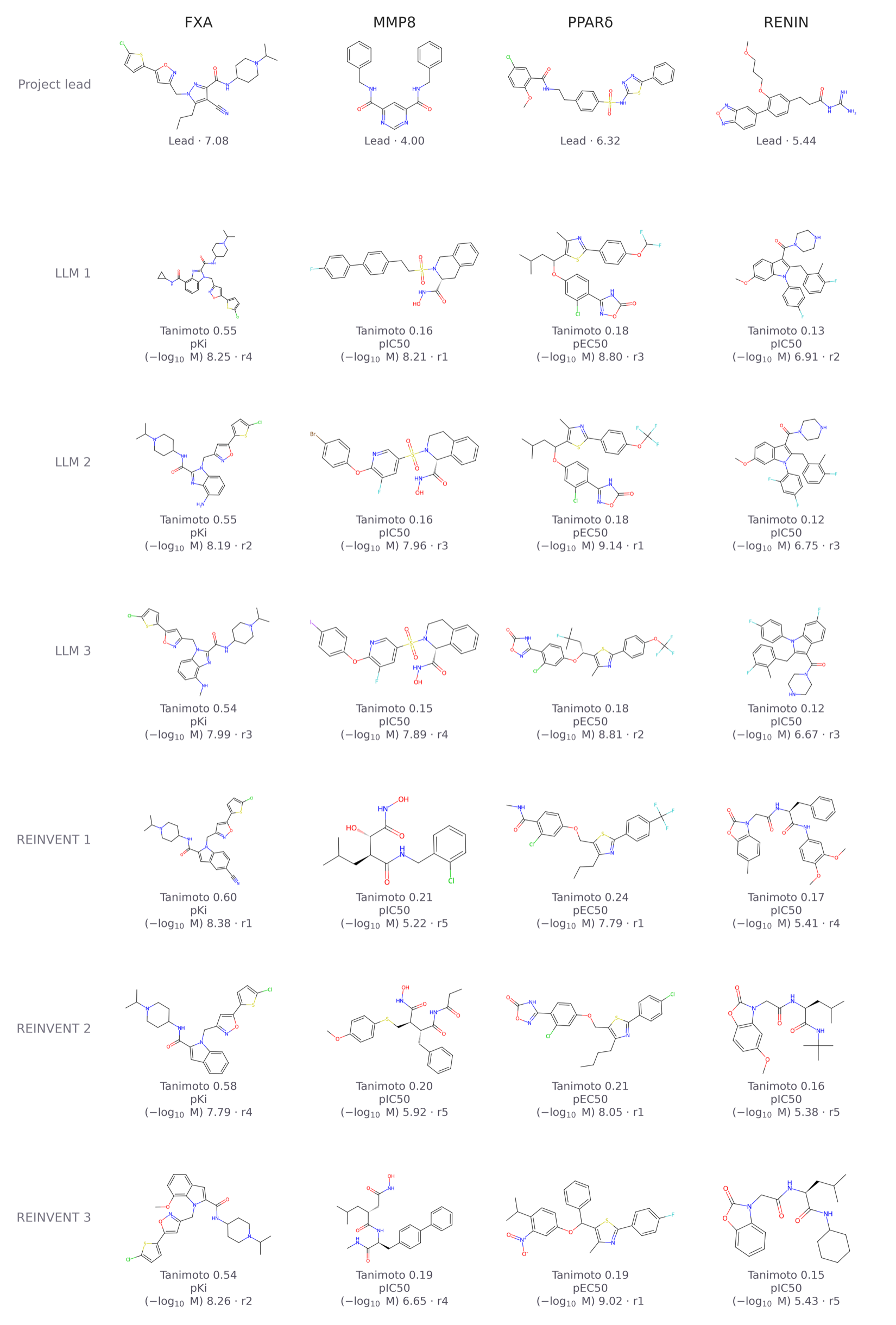}
  \caption{\textbf{Best held-out compounds and top-ranked LLM and REINVENT
    proposals for the Sanofi projects.} Best measured held-out molecule plus the
    top three LLM and top three REINVENT proposals per target.}
  \label{fig:sanofi-best}
\end{figure}

\begin{figure}[p]
  \centering
  \includegraphics[width=\linewidth, height=0.9\textheight, keepaspectratio]{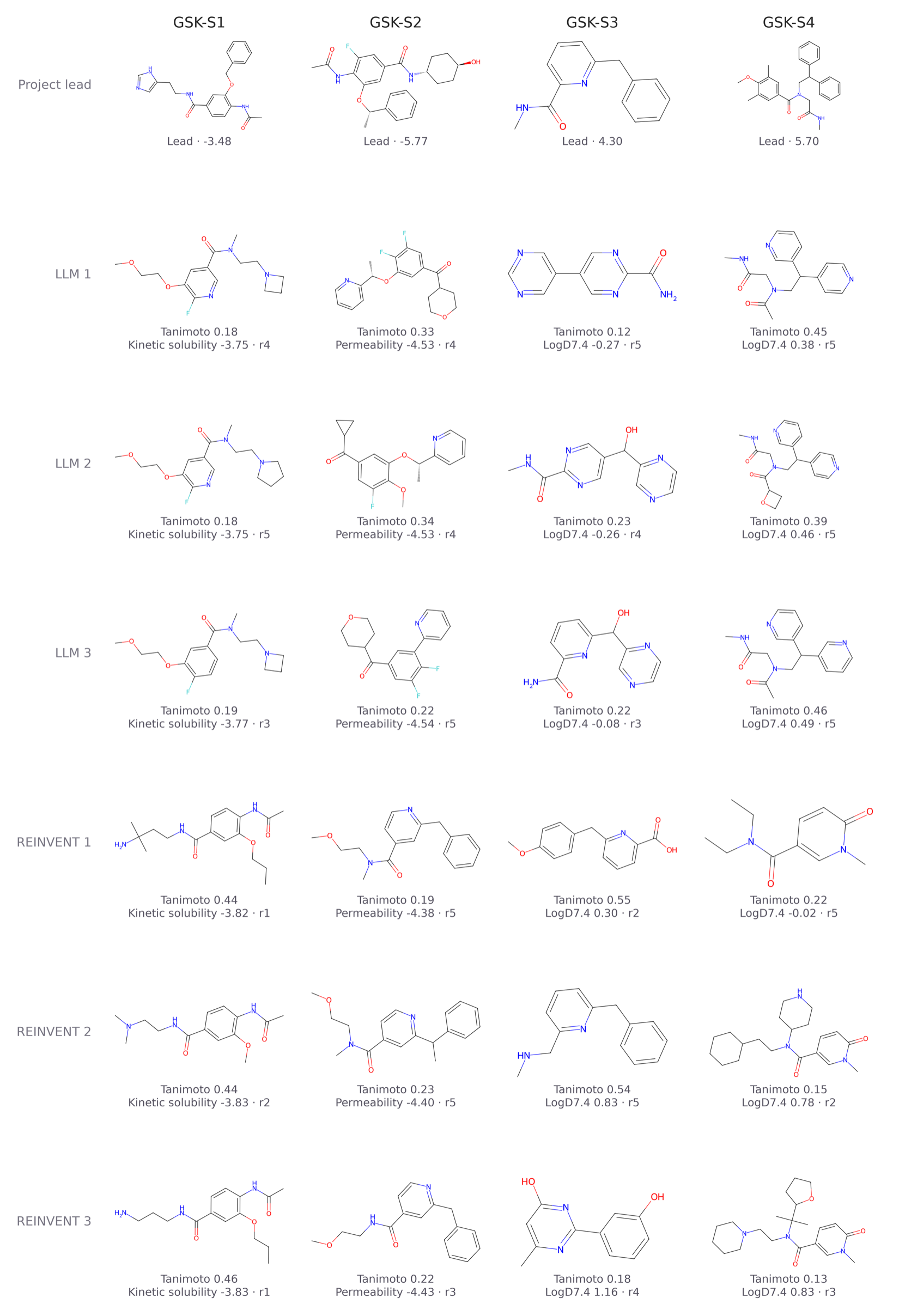}
  \caption{\textbf{Top five LLM and REINVENT proposals for each GSK project.}
    Columns correspond to GSK-S1--S4; rows show ranks 1--5 for LLM followed by
    ranks 1--5 for REINVENT. Structures are ranked in the favourable direction
    of the model-predicted primary endpoint across rounds 1--5 after
    canonical-SMILES deduplication. The predicted endpoint value and originating
    round are below each molecule.}
  \label{fig:gsk-top5}
\end{figure}

\begin{figure}[p]
  \centering
  \includegraphics[width=\linewidth, height=0.9\textheight, keepaspectratio]{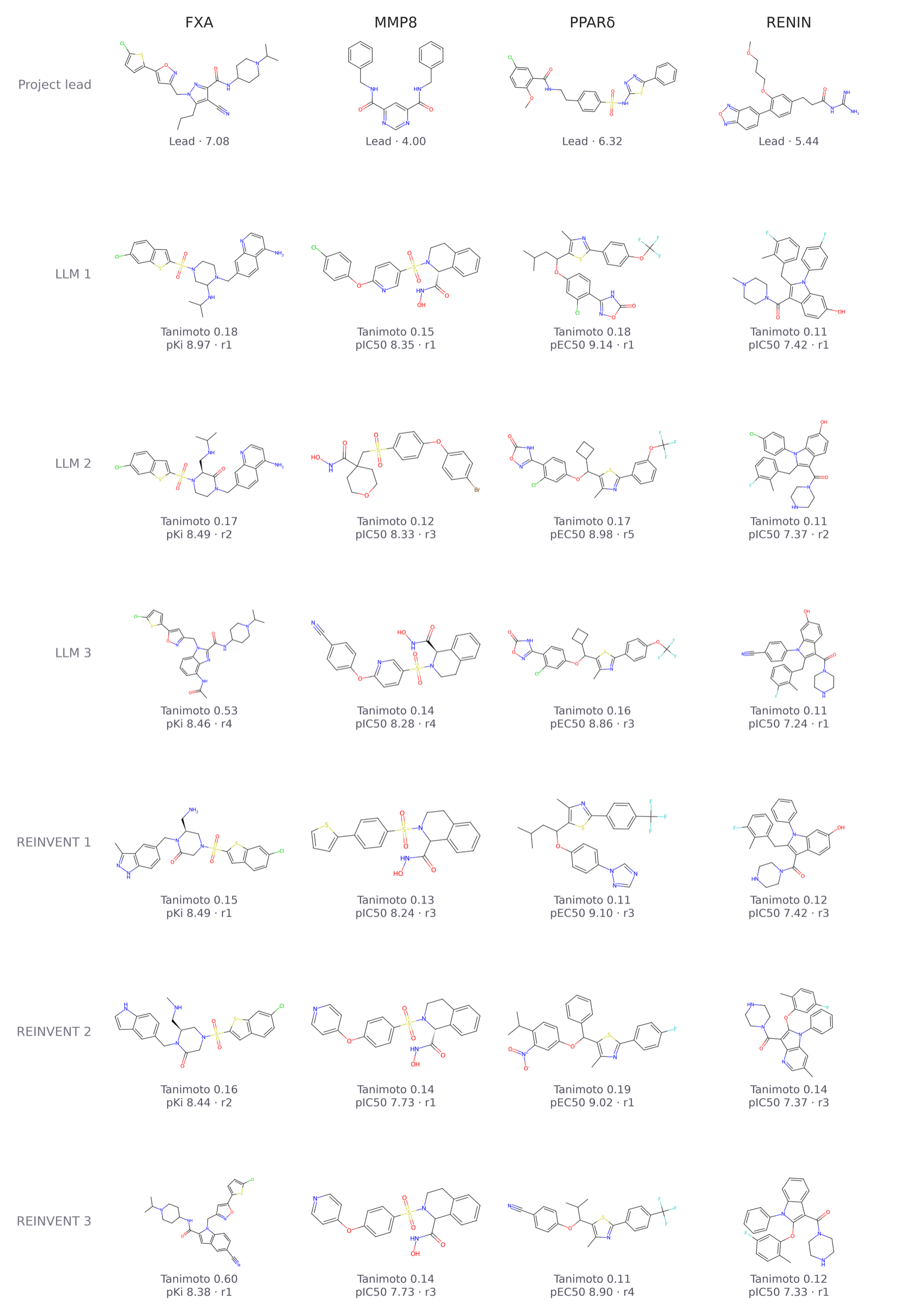}
  \caption{\textbf{Top five LLM and REINVENT proposals for each Sanofi project.}
    Columns correspond to FXA, MMP8, \PPARd{} and RENIN; rows show ranks 1--5
    for each route. Ranking uses the project model for \pKi{}, \pICfifty{} or
    \pECfifty{} prediction in the favorable direction across rounds 1--5 after
    canonical-SMILES deduplication. Values and originating rounds are printed
    below each structure.}
  \label{fig:sanofi-top5}
\end{figure}

\begin{figure}[p]
  \centering
  \includegraphics[width=\linewidth, height=0.9\textheight, keepaspectratio]{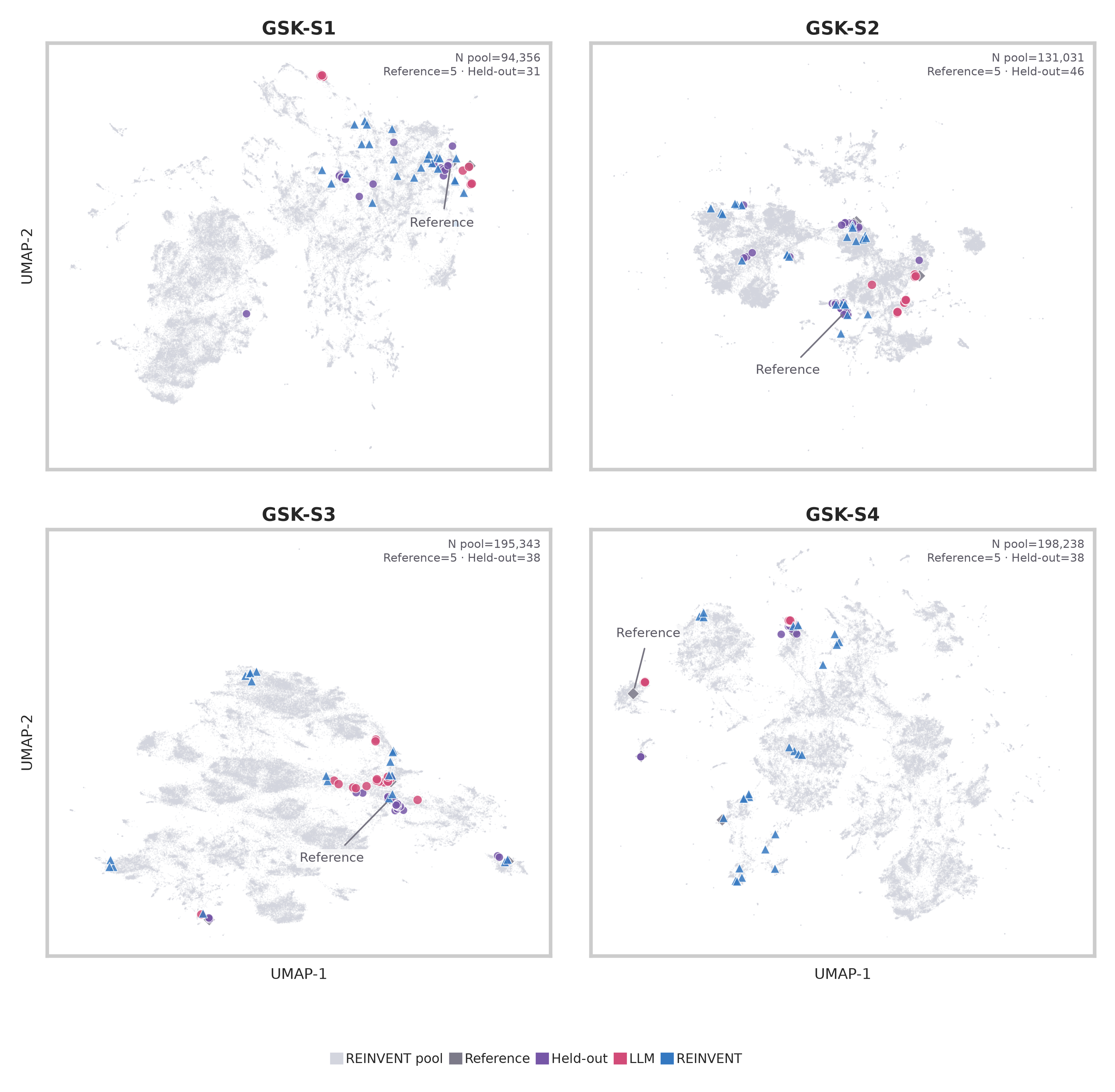}
  \caption{\textbf{UMAP chemical space coverage for the four GSK projects.}
    Light-gray circles represent unique structures from the internal REINVENT
    pools, purple diamonds the five project Reference compounds, green circles
    Held-out compounds, red circles final LLM proposals and blue triangles final
    REINVENT proposals. Embeddings use 2,048-bit ECFP4 fingerprints and Jaccard
    distance. Each project is embedded independently. UMAP was computed with
    umap-learn \cite{mcinnes2018umap}.}
  \label{fig:umap-gsk}
\end{figure}

\begin{figure}[p]
  \centering
  \includegraphics[width=\linewidth, height=0.9\textheight, keepaspectratio]{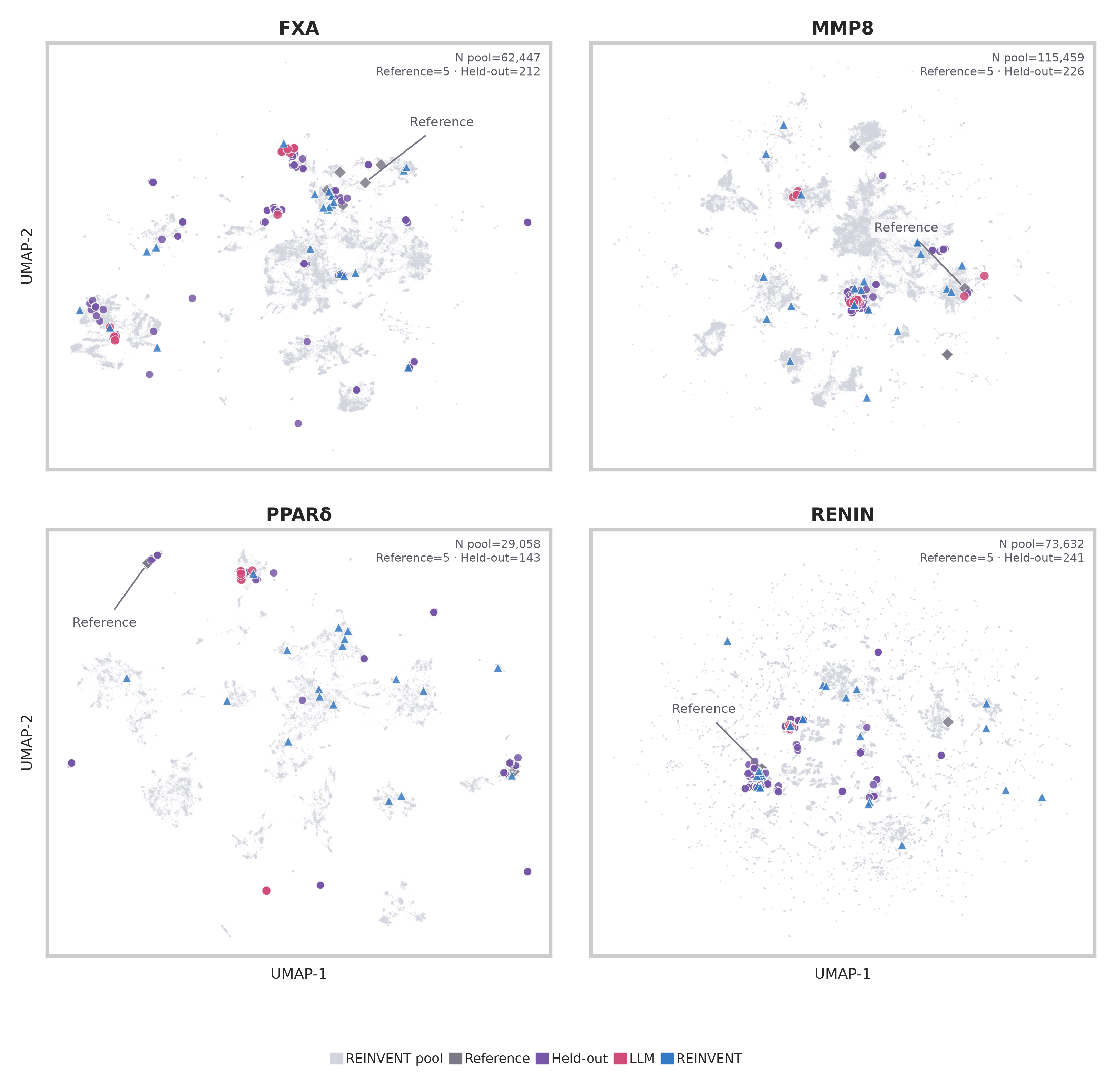}
  \caption{\textbf{UMAP chemical space coverage for the four Sanofi projects.}
    Light-gray circles represent unique structures from the internal REINVENT
    pools, purple diamonds the five project Reference compounds, green circles
    Held-out compounds, red circles final LLM proposals and blue triangles final
    REINVENT proposals. Embeddings use 2,048-bit ECFP4 fingerprints and Jaccard
    distance. Each project is embedded independently. UMAP was computed with
    umap-learn \cite{mcinnes2018umap}.}
  \label{fig:umap-sanofi}
\end{figure}

\begin{figure}[p]
  \centering
  \includegraphics[width=\linewidth, height=0.9\textheight, keepaspectratio]{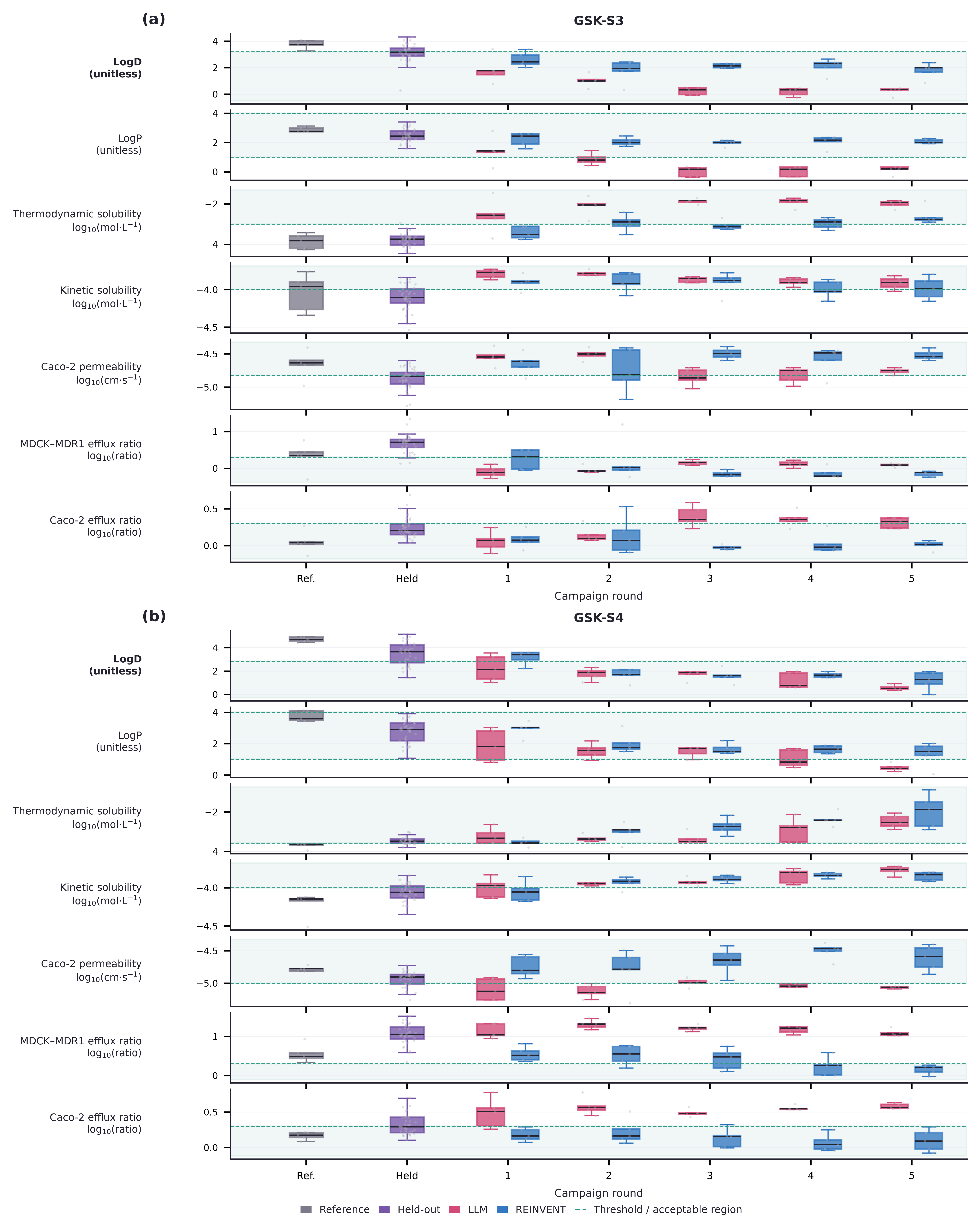}
  \caption{\textbf{Property distributions during rounds 1--5 of the GSK-S3 and
    GSK-S4 campaigns.} Gray denotes the five shown Reference compounds, green
    for the published Held-out compounds, red LLM proposals and blue final
    agent-selected REINVENT proposals. Points represent individual molecules;
    boxes show the median and interquartile range, with whiskers extending 1.5
    times the interquartile range. Each row identifies the endpoint and unit.
    Dashed green lines mark project-specific criteria where available and
    generic ADMET context ranges otherwise; pale-green shading shows the
    corresponding favorable region. Reference and Held-out values are study
    measurements, whereas proposal values are model predictions.}
  \label{fig:gsk34-properties}
\end{figure}

\begin{figure}[p]
  \centering
  \includegraphics[width=\linewidth, height=0.9\textheight, keepaspectratio]{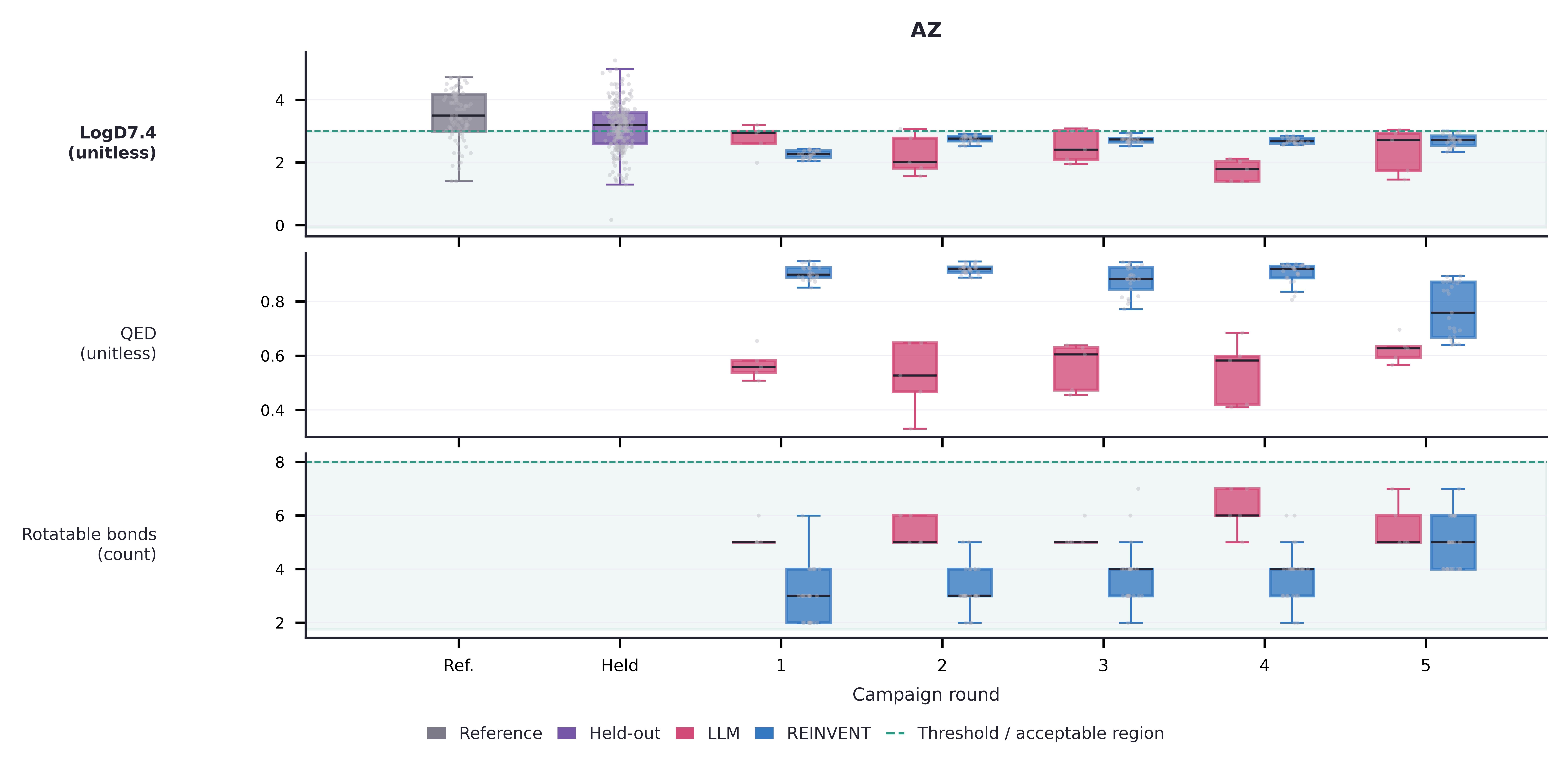}
  \caption{\textbf{Property distributions during rounds 1--5 of the AstraZeneca
    $\log D_{7.4}$ campaign.} Gray denotes the five shown Reference compounds, green
    for the published Held-out compounds, red LLM proposals and blue final
    agent-selected REINVENT proposals. Points represent individual molecules; boxes show the median and interquartile range, with whiskers extending 1.5 times the interquartile
    range. Each row identifies the endpoint and unit. Reference and Held-out
    values are study measurements and are available for $\log D_{7.4}$ only; QED and
    rotatable-bond counts are RDKit descriptors computed on the proposals.
    Dashed green lines mark the campaign criteria, and pale-green shading shows
    the corresponding favorable region.}
  \label{fig:az-properties}
\end{figure}

\begin{figure}[p]
  \centering
  \includegraphics[width=\linewidth, height=0.9\textheight, keepaspectratio]{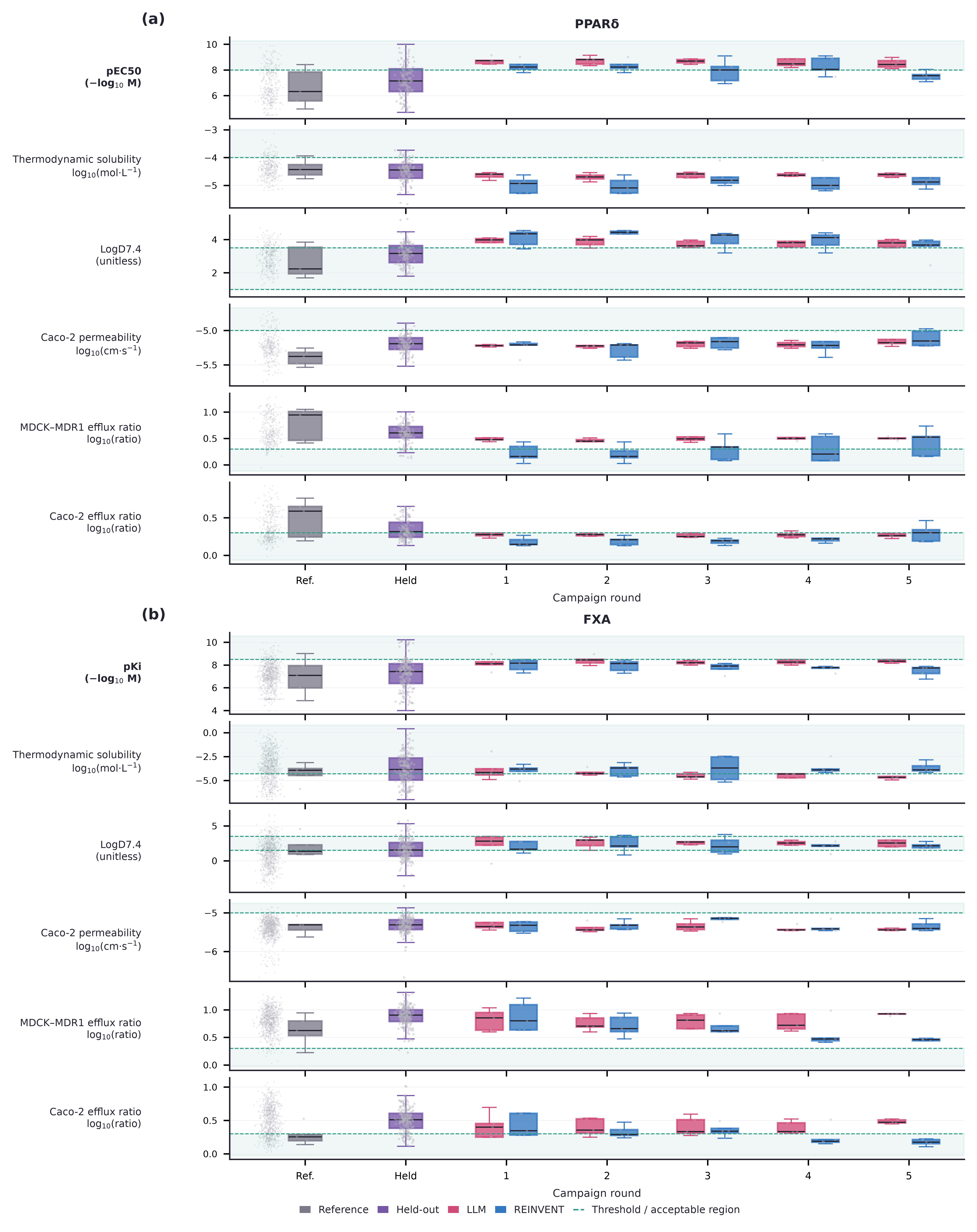}
  \caption{\textbf{Property distributions during rounds 1--5 of the Sanofi PPARd
    and FXA campaigns.} Gray boxes denote the retrospective five-compound
    Reference samples used in Figure~\ref{fig:main-quality}, pale-gray points
    the complete pre-cutoff project-model training set, green boxes the future
    Held-out compounds, red boxes LLM proposals and blue boxes final
    agent-selected REINVENT proposals. The five-compound Reference samples
    provide visualization context and were not the parents used in the
    campaigns. Points represent molecules; boxes show medians and interquartile
    ranges, with 1.5-interquartile-range whiskers. Historical primary-potency
    values---and FXA solubility and $\log D_{7.4}$---are experimental measurements.
    Generic ADMET rows contain model predictions, while TPSA and molecular
    weight are RDKit descriptors. Dashed green lines show project criteria or
    generic contextual ranges, and pale-green shading marks the corresponding
    favorable region.}
  \label{fig:ppard-fxa-properties}
\end{figure}

\begin{figure}[p]
  \centering
  \includegraphics[width=\linewidth, height=0.9\textheight, keepaspectratio]{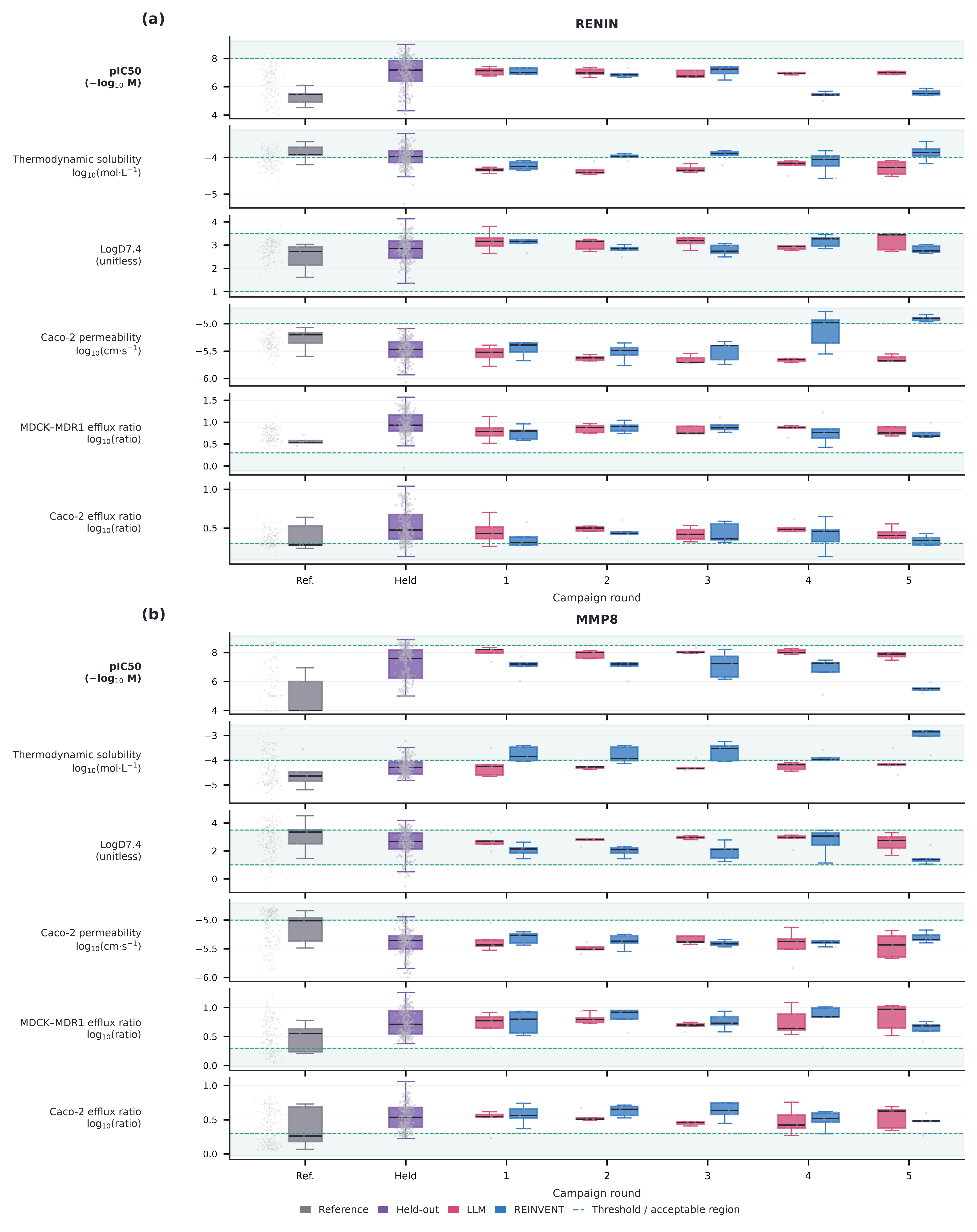}
  \caption{\textbf{Property distributions during rounds 1--5 of the four Sanofi
    RENIN and MMP8 campaigns.} Gray boxes denote the retrospective five-compound
    Reference samples used in Figure~\ref{fig:main-quality}, pale-gray points
    the complete pre-cutoff project-model training set, green boxes the future
    Held-out compounds, red boxes LLM proposals and blue boxes final
    agent-selected REINVENT proposals. The five-compound Reference samples
    provide visualization context and were not the parents used in the
    campaigns. Points represent molecules; boxes show medians and interquartile
    ranges, with 1.5-interquartile-range whiskers. Historical primary-potency
    values---and FXA solubility and $\log D_{7.4}$---are experimental measurements.
    Generic ADMET rows contain model predictions, while TPSA and molecular
    weight are RDKit descriptors. Dashed green lines show project criteria or
    generic contextual ranges, and pale-green shading marks the corresponding
    favorable region.}
  \label{fig:renin-mmp8-properties}
\end{figure}

\begin{figure}[htbp]
  \centering
  \includegraphics[width=\linewidth]{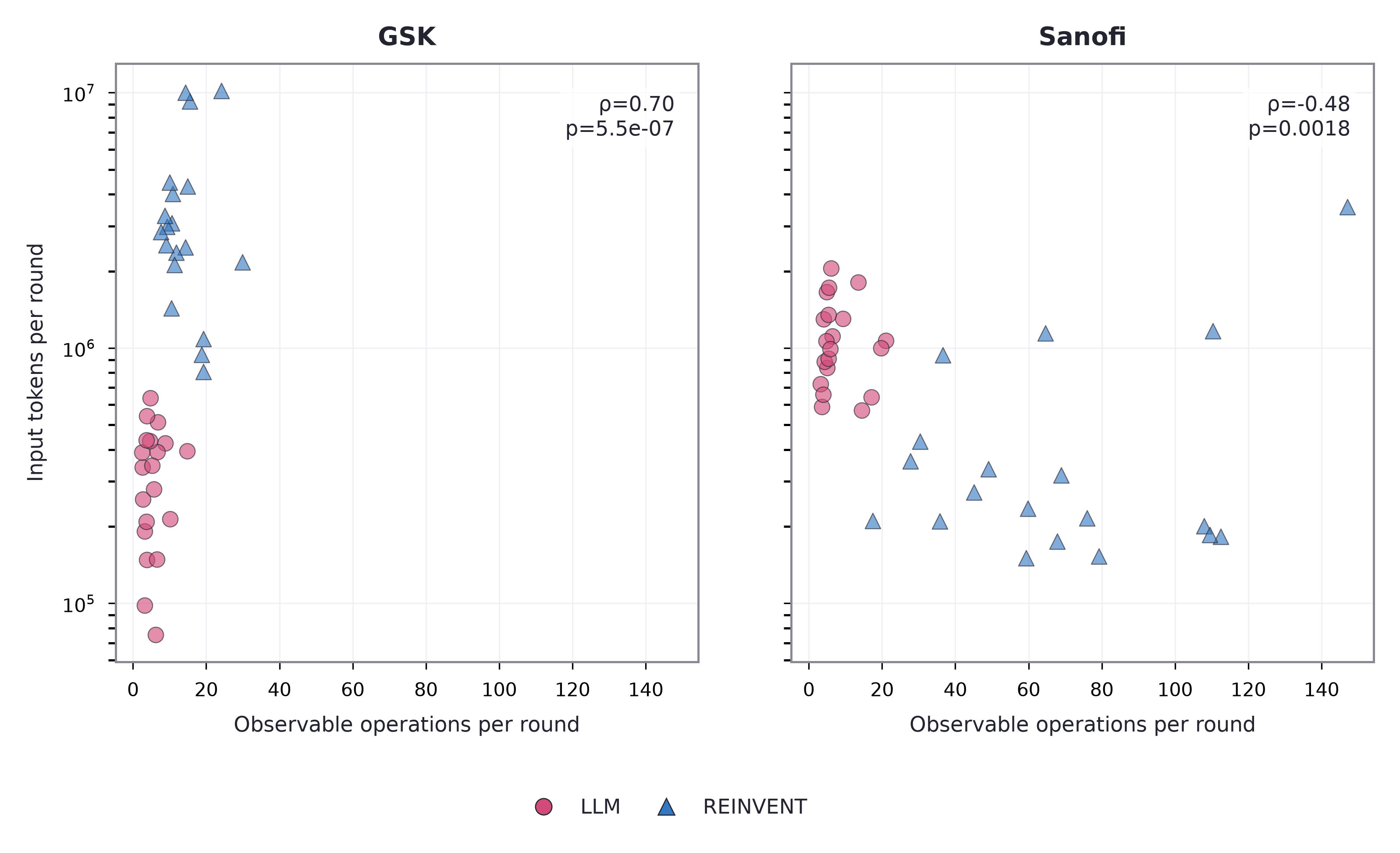}
  \caption{\textbf{Input-token volume versus observable agent operations during
    rounds 1--5.} LLM and REINVENT values include model-context tokens and
    agent-visible operations; internal REINVENT scorer calls are excluded.}
  \label{fig:tokens}
\end{figure}

%%%%%%%%%%%%%%%%%%%%%%%%%%%%%%%%%%%%%%%%%%%%%%%%%%%%%%%%%%%%

\newpage
\section*{NeurIPS Paper Checklist}

\begin{enumerate}

\item {\bf Claims}
    \item[] Question: Do the main claims made in the abstract and introduction accurately reflect the paper's contributions and scope?
    \item[] Answer: \answerYes{}
    \item[] Justification: The three contributions listed in Section~\ref{sec:introduction} map directly onto the reported results: the versioned executable objective shared by both generation routes (Section~\ref{sec:objective}), the campaign-level evaluation on nine industry-derived tasks (Sections~\ref{sec:quality}--\ref{sec:operations}), and the analysis of evidence-driven adaptation (Section~\ref{sec:operations} and Section~\ref{sec:adaptation}). The abstract states that attainment was governed primarily by predictor calibration rather than generation route, which is the conclusion supported by Section~\ref{sec:attainment}.
    \item[] Guidelines:
    \begin{itemize}
        \item The answer \answerNA{} means that the abstract and introduction do not include the claims made in the paper.
        \item The abstract and/or introduction should clearly state the claims made, including the contributions made in the paper and important assumptions and limitations. A \answerNo{} or \answerNA{} answer to this question will not be perceived well by the reviewers. 
        \item The claims made should match theoretical and experimental results, and reflect how much the results can be expected to generalize to other settings. 
        \item It is fine to include aspirational goals as motivation as long as it is clear that these goals are not attained by the paper. 
    \end{itemize}

\item {\bf Limitations}
    \item[] Question: Does the paper discuss the limitations of the work performed by the authors?
    \item[] Answer: \answerYes{}
    \item[] Justification: Section~\ref{sec:limitations} is a dedicated Limitations section. It states that all proposal endpoints are predictions,
    so historical measurements represent a different evidence level; that each
    condition used one reasoning backend and one trajectory; and that temporal splits cannot exclude pretraining
    exposure. 
    \item[] Guidelines:
    \begin{itemize}
        \item The answer \answerNA{} means that the paper has no limitation while the answer \answerNo{} means that the paper has limitations, but those are not discussed in the paper. 
        \item The authors are encouraged to create a separate ``Limitations'' section in their paper.
        \item The paper should point out any strong assumptions and how robust the results are to violations of these assumptions (e.g., independence assumptions, noiseless settings, model well-specification, asymptotic approximations only holding locally). The authors should reflect on how these assumptions might be violated in practice and what the implications would be.
        \item The authors should reflect on the scope of the claims made, e.g., if the approach was only tested on a few datasets or with a few runs. In general, empirical results often depend on implicit assumptions, which should be articulated.
        \item The authors should reflect on the factors that influence the performance of the approach. For example, a facial recognition algorithm may perform poorly when image resolution is low or images are taken in low lighting. Or a speech-to-text system might not be used reliably to provide closed captions for online lectures because it fails to handle technical jargon.
        \item The authors should discuss the computational efficiency of the proposed algorithms and how they scale with dataset size.
        \item If applicable, the authors should discuss possible limitations of their approach to address problems of privacy and fairness.
        \item While the authors might fear that complete honesty about limitations might be used by reviewers as grounds for rejection, a worse outcome might be that reviewers discover limitations that aren't acknowledged in the paper. The authors should use their best judgment and recognize that individual actions in favor of transparency play an important role in developing norms that preserve the integrity of the community. Reviewers will be specifically instructed to not penalize honesty concerning limitations.
    \end{itemize}

\item {\bf Theory assumptions and proofs}
    \item[] Question: For each theoretical result, does the paper provide the full set of assumptions and a complete (and correct) proof?
    \item[] Answer: \answerNA{}
    \item[] Justification: The paper contains no theoretical results. Equations~\eqref{eq:desirability} and~\eqref{eq:geometric-mean} are definitions of the endpoint desirability transform and the weighted Derringer--Suich aggregation adopted from prior work \cite{derringer1980}, not claims requiring proof.
    \item[] Guidelines:
    \begin{itemize}
        \item The answer \answerNA{} means that the paper does not include theoretical results. 
        \item All the theorems, formulas, and proofs in the paper should be numbered and cross-referenced.
        \item All assumptions should be clearly stated or referenced in the statement of any theorems.
        \item The proofs can either appear in the main paper or the supplemental material, but if they appear in the supplemental material, the authors are encouraged to provide a short proof sketch to provide intuition. 
        \item Inversely, any informal proof provided in the core of the paper should be complemented by formal proofs provided in appendix or supplemental material.
        \item Theorems and Lemmas that the proof relies upon should be properly referenced. 
    \end{itemize}

    \item {\bf Experimental result reproducibility}
    \item[] Question: Does the paper fully disclose all the information needed to reproduce the main experimental results of the paper to the extent that it affects the main claims and/or conclusions of the paper (regardless of whether the code and data are provided or not)?
    \item[] Answer: \answerYes{}
    \item[] Justification: Section~\ref{sec:agent} specifies the agent architecture, the TTP schema and the tool loop; Section~\ref{sec:datasets} specifies dataset curation, standardization and the temporal cutoffs; Section~\ref{sec:services} specifies every model architecture and hyperparameter; and Section~\ref{sec:validation} specifies the campaign protocol. Table~\ref{tab:services} lists deployed service versions, Table~\ref{tab:objectives} the per-project objectives and guardrails, and Table~\ref{tab:reinvent-params} the exact per-invocation REINVENT parameters. The reasoning backend is identified in Section~\ref{sec:system-prompt}, which also notes that provider-side sampling defaults were used and no seed was fixed.
    \item[] Guidelines:
    \begin{itemize}
        \item The answer \answerNA{} means that the paper does not include experiments.
        \item If the paper includes experiments, a \answerNo{} answer to this question will not be perceived well by the reviewers: Making the paper reproducible is important, regardless of whether the code and data are provided or not.
        \item If the contribution is a dataset and\slash or model, the authors should describe the steps taken to make their results reproducible or verifiable. 
        \item Depending on the contribution, reproducibility can be accomplished in various ways. For example, if the contribution is a novel architecture, describing the architecture fully might suffice, or if the contribution is a specific model and empirical evaluation, it may be necessary to either make it possible for others to replicate the model with the same dataset, or provide access to the model. In general. releasing code and data is often one good way to accomplish this, but reproducibility can also be provided via detailed instructions for how to replicate the results, access to a hosted model (e.g., in the case of a large language model), releasing of a model checkpoint, or other means that are appropriate to the research performed.
        \item While NeurIPS does not require releasing code, the conference does require all submissions to provide some reasonable avenue for reproducibility, which may depend on the nature of the contribution. For example
        \begin{enumerate}
            \item If the contribution is primarily a new algorithm, the paper should make it clear how to reproduce that algorithm.
            \item If the contribution is primarily a new model architecture, the paper should describe the architecture clearly and fully.
            \item If the contribution is a new model (e.g., a large language model), then there should either be a way to access this model for reproducing the results or a way to reproduce the model (e.g., with an open-source dataset or instructions for how to construct the dataset).
            \item We recognize that reproducibility may be tricky in some cases, in which case authors are welcome to describe the particular way they provide for reproducibility. In the case of closed-source models, it may be that access to the model is limited in some way (e.g., to registered users), but it should be possible for other researchers to have some path to reproducing or verifying the results.
        \end{enumerate}
    \end{itemize}

\item {\bf Open access to data and code}
    \item[] Question: Does the paper provide open access to the data and code, with sufficient instructions to faithfully reproduce the main experimental results, as described in supplemental material?
    \item[] Answer: \answerYes{}
    \item[] Justification: The MAGI agent, its MCP server and the global ADMET
    ensemble will be released under an open-source licence upon publication,
    with instructions covering environment setup, service contracts and the
    commands needed to replay a campaign. The RetroDMTA projects are third-party
    assets released under \cite{mas2026retrodmta} with an accompanying code DOI,
    and the GSK series are extracted from published articles
    \cite{gsk046series, nmethylpyridineseries, messenger2021template,
    taylor2021del}. The AstraZeneca campaign data are not releasable; results depending on them are therefore not reproducible from
    released artefacts alone.
    \item[] Guidelines:
    \begin{itemize}
        \item The answer \answerNA{} means that paper does not include experiments requiring code.
        \item Please see the NeurIPS code and data submission guidelines (\url{https://neurips.cc/public/guides/CodeSubmissionPolicy}) for more details.
        \item While we encourage the release of code and data, we understand that this might not be possible, so \answerNo{} is an acceptable answer. Papers cannot be rejected simply for not including code, unless this is central to the contribution (e.g., for a new open-source benchmark).
        \item The instructions should contain the exact command and environment needed to run to reproduce the results. See the NeurIPS code and data submission guidelines (\url{https://neurips.cc/public/guides/CodeSubmissionPolicy}) for more details.
        \item The authors should provide instructions on data access and preparation, including how to access the raw data, preprocessed data, intermediate data, and generated data, etc.
        \item The authors should provide scripts to reproduce all experimental results for the new proposed method and baselines. If only a subset of experiments are reproducible, they should state which ones are omitted from the script and why.
        \item At submission time, to preserve anonymity, the authors should release anonymized versions (if applicable).
        \item Providing as much information as possible in supplemental material (appended to the paper) is recommended, but including URLs to data and code is permitted.
    \end{itemize}

\item {\bf Experimental setting/details}
    \item[] Question: Does the paper specify all the training and test details (e.g., data splits, hyperparameters, how they were chosen, type of optimizer) necessary to understand the results?
    \item[] Answer: \answerYes{}
    \item[] Justification: Section~\ref{sec:global-admet} gives the ADMET model architectures, loss, task weighting, epoch cap and early-stopping patience, and the five shared Butina folds; Section~\ref{sec:project-models} gives the project-model recipe and temporal restriction; Section~\ref{sec:boltz2} gives the recycling and diffusion sampling settings. Splits and cohort sizes are in Table~\ref{tab:endpoint-composition} and Section~\ref{sec:retrodmta}, and per-invocation generation hyperparameters in Table~\ref{tab:reinvent-params}.
    \item[] Guidelines:
    \begin{itemize}
        \item The answer \answerNA{} means that the paper does not include experiments.
        \item The experimental setting should be presented in the core of the paper to a level of detail that is necessary to appreciate the results and make sense of them.
        \item The full details can be provided either with the code, in appendix, or as supplemental material.
    \end{itemize}

\item {\bf Experiment statistical significance}
    \item[] Question: Does the paper report error bars suitably and correctly defined or other appropriate information about the statistical significance of the experiments?
    \item[] Answer: \answerYes{}
    \item[] Justification: Section~\ref{sec:statistics} defines the experimental unit (project) and the repeated measure (round), and specifies two-sided Wilcoxon signed-rank tests over four paired projects per organization with Holm adjustment for Figure~\ref{fig:main-quality}a. Box plots report medians, interquartile ranges and 1.5-IQR whiskers, and leakage metrics use bootstrap 95\% intervals (Section~\ref{sec:molecular-quality}). Section~\ref{sec:statistics} explicitly labels the within-project molecule-level Mann--Whitney comparisons as exploratory because analogues remain chemically dependent.
    \item[] Guidelines:
    \begin{itemize}
        \item The answer \answerNA{} means that the paper does not include experiments.
        \item The authors should answer \answerYes{} if the results are accompanied by error bars, confidence intervals, or statistical significance tests, at least for the experiments that support the main claims of the paper.
        \item The factors of variability that the error bars are capturing should be clearly stated (for example, train/test split, initialization, random drawing of some parameter, or overall run with given experimental conditions).
        \item The method for calculating the error bars should be explained (closed form formula, call to a library function, bootstrap, etc.)
        \item The assumptions made should be given (e.g., Normally distributed errors).
        \item It should be clear whether the error bar is the standard deviation or the standard error of the mean.
        \item It is OK to report 1-sigma error bars, but one should state it. The authors should preferably report a 2-sigma error bar than state that they have a 96\% CI, if the hypothesis of Normality of errors is not verified.
        \item For asymmetric distributions, the authors should be careful not to show in tables or figures symmetric error bars that would yield results that are out of range (e.g., negative error rates).
        \item If error bars are reported in tables or plots, the authors should explain in the text how they were calculated and reference the corresponding figures or tables in the text.
    \end{itemize}

\item {\bf Experiments compute resources}
    \item[] Question: For each experiment, does the paper provide sufficient information on the computer resources (type of compute workers, memory, time of execution) needed to reproduce the experiments?
    \item[] Answer: \answerYes{}
    \item[] Justification: Section~\ref{sec:compute} reports the hardware (AMD EPYC 9575F CPUs, AMD Instinct MI355X accelerators, ROCm 7.0), states that REINVENT and Boltz-2 occupied separate GPUs and that the remaining services were CPU-parallelized. Generation workloads are in Tables~\ref{tab:pool} and~\ref{tab:reinvent-params}, and Figure~\ref{fig:tokens} reports input-token volume against observable operations.
    \item[] Guidelines:
    \begin{itemize}
        \item The answer \answerNA{} means that the paper does not include experiments.
        \item The paper should indicate the type of compute workers CPU or GPU, internal cluster, or cloud provider, including relevant memory and storage.
        \item The paper should provide the amount of compute required for each of the individual experimental runs as well as estimate the total compute. 
        \item The paper should disclose whether the full research project required more compute than the experiments reported in the paper (e.g., preliminary or failed experiments that didn't make it into the paper). 
    \end{itemize}
    
\item {\bf Code of ethics}
    \item[] Question: Does the research conducted in the paper conform, in every respect, with the NeurIPS Code of Ethics \url{https://neurips.cc/public/EthicsGuidelines}?
    \item[] Answer: \answerYes{}
    \item[] Justification: The research uses published medicinal-chemistry literature and de-identified industrial project data, releases no personally identifying information about the blinded evaluators (Section~\ref{sec:ethics}), and discusses the dual-use profile of molecular design in Section~\ref{sec:responsible-release}. Anonymity is preserved throughout the submission.
    \item[] Guidelines:
    \begin{itemize}
        \item The answer \answerNA{} means that the authors have not reviewed the NeurIPS Code of Ethics.
        \item If the authors answer \answerNo, they should explain the special circumstances that require a deviation from the Code of Ethics.
        \item The authors should make sure to preserve anonymity (e.g., if there is a special consideration due to laws or regulations in their jurisdiction).
    \end{itemize}

\item {\bf Broader impacts}
    \item[] Question: Does the paper discuss both potential positive societal impacts and negative societal impacts of the work performed?
    \item[] Answer: \answerYes{}
    \item[] Justification: Section~\ref{sec:responsible-release} states that molecular design systems have potential dual-use applications and may generate unsafe, reactive or otherwise inappropriate structures. The positive framing, accelerating lead optimization as an adaptive coordination layer over established models rather than replacing them, is developed in Sections~\ref{sec:discussion} and~\ref{sec:conclusion}.
    \item[] Guidelines:
    \begin{itemize}
        \item The answer \answerNA{} means that there is no societal impact of the work performed.
        \item If the authors answer \answerNA{} or \answerNo, they should explain why their work has no societal impact or why the paper does not address societal impact.
        \item Examples of negative societal impacts include potential malicious or unintended uses (e.g., disinformation, generating fake profiles, surveillance), fairness considerations (e.g., deployment of technologies that could make decisions that unfairly impact specific groups), privacy considerations, and security considerations.
        \item The conference expects that many papers will be foundational research and not tied to particular applications, let alone deployments. However, if there is a direct path to any negative applications, the authors should point it out. For example, it is legitimate to point out that an improvement in the quality of generative models could be used to generate Deepfakes for disinformation. On the other hand, it is not needed to point out that a generic algorithm for optimizing neural networks could enable people to train models that generate Deepfakes faster.
        \item The authors should consider possible harms that could arise when the technology is being used as intended and functioning correctly, harms that could arise when the technology is being used as intended but gives incorrect results, and harms following from (intentional or unintentional) misuse of the technology.
        \item If there are negative societal impacts, the authors could also discuss possible mitigation strategies (e.g., gated release of models, providing defenses in addition to attacks, mechanisms for monitoring misuse, mechanisms to monitor how a system learns from feedback over time, improving the efficiency and accessibility of ML).
    \end{itemize}
    
\item {\bf Safeguards}
    \item[] Question: Does the paper describe safeguards that have been put in place for responsible release of data or models that have a high risk for misuse (e.g., pre-trained language models, image generators, or scraped datasets)?
    \item[] Answer: \answerYes{}
    \item[] Justification: Section~\ref{sec:responsible-release} states that MAGI outputs are predictions rather than experimentally validated compounds and require expert review. The system is an orchestration layer over existing published generators and predictors and introduces no new generative capability beyond REINVENT 4 and the configured reasoning backend, and the industrial project data are de-identified.
    \item[] Guidelines:
    \begin{itemize}
        \item The answer \answerNA{} means that the paper poses no such risks.
        \item Released models that have a high risk for misuse or dual-use should be released with necessary safeguards to allow for controlled use of the model, for example by requiring that users adhere to usage guidelines or restrictions to access the model or implementing safety filters. 
        \item Datasets that have been scraped from the Internet could pose safety risks. The authors should describe how they avoided releasing unsafe images.
        \item We recognize that providing effective safeguards is challenging, and many papers do not require this, but we encourage authors to take this into account and make a best faith effort.
    \end{itemize}

\item {\bf Licenses for existing assets}
    \item[] Question: Are the creators or original owners of assets (e.g., code, data, models), used in the paper, properly credited and are the license and terms of use explicitly mentioned and properly respected?
    \item[] Answer: \answerYes{}
    \item[] Justification: Every external asset is cited with the deployed version in Table~\ref{tab:services}: REINVENT 4 \cite{loeffler2024reinvent4}, Boltz-2 \cite{passaro2025boltz2}, PoseBusters \cite{buttenschoen2024posebusters}, RDKit \cite{rdkit}, XGBoost \cite{chen2016xgboost}, Chemprop \cite{yang2019chemprop}, CheMeleon \cite{burns2026chemeleon} and UMAP \cite{mcinnes2018umap}. Data sources are credited as OCHEM \cite{sushko2011ochem}, ChEMBL \cite{zdrazil2024chembl}, the PDB \cite{burley2019rcsb}, UniProt \cite{uniprot2023}, the RetroDMTA benchmark \cite{mas2026retrodmta} and the four published GSK series. % AUTHORS: consider naming the specific licenses, which the guidelines request but the manuscript does not currently state.
    \item[] Guidelines:
    \begin{itemize}
        \item The answer \answerNA{} means that the paper does not use existing assets.
        \item The authors should cite the original paper that produced the code package or dataset.
        \item The authors should state which version of the asset is used and, if possible, include a URL.
        \item The name of the license (e.g., CC-BY 4.0) should be included for each asset.
        \item For scraped data from a particular source (e.g., website), the copyright and terms of service of that source should be provided.
        \item If assets are released, the license, copyright information, and terms of use in the package should be provided. For popular datasets, \url{paperswithcode.com/datasets} has curated licenses for some datasets. Their licensing guide can help determine the license of a dataset.
        \item For existing datasets that are re-packaged, both the original license and the license of the derived asset (if it has changed) should be provided.
        \item If this information is not available online, the authors are encouraged to reach out to the asset's creators.
    \end{itemize}

\item {\bf New assets}
    \item[] Question: Are new assets introduced in the paper well documented and is the documentation provided alongside the assets?
    \item[] Answer: \answerYes{}
    \item[] Justification: The paper introduces the MAGI agent, its MCP server,
    the external REST scoring component for REINVENT 4, the global ADMET
    ensemble and the interaction profiler. These are described in
    Sections~\ref{sec:agent} and~\ref{sec:services}. Code and the trained global
    ADMET ensemble will be released under an open-source licence upon
    publication, with documentation covering installation, service contracts,
    applicability-domain caveats and known limitations. 
    \item[] Guidelines:
    \begin{itemize}
        \item The answer \answerNA{} means that the paper does not release new assets.
        \item Researchers should communicate the details of the dataset\slash code\slash model as part of their submissions via structured templates. This includes details about training, license, limitations, etc. 
        \item The paper should discuss whether and how consent was obtained from people whose asset is used.
        \item At submission time, remember to anonymize your assets (if applicable). You can either create an anonymized URL or include an anonymized zip file.
    \end{itemize}

\item {\bf Crowdsourcing and research with human subjects}
    \item[] Question: For crowdsourcing experiments and research with human subjects, does the paper include the full text of instructions given to participants and screenshots, if applicable, as well as details about compensation (if any)? 
    \item[] Answer: \answerYes{} 
    \item[] Justification: The blinded chemist evaluation involves human
    subjects. Section~\ref{sec:medchem-eval} describes the instrument in full:
    28 questions across three case types, the exact response format required for
    each, and the primary analyses. Participants were four practising medicinal
    chemists or computational chemists; the exercise formed part
    of their normal work and no separate compensation was offered.
    Section~\ref{sec:ethics} states that no personally identifying participant
    information is released.

    \item[] Guidelines:
    \begin{itemize}
        \item The answer \answerNA{} means that the paper does not involve crowdsourcing nor research with human subjects.
        \item Including this information in the supplemental material is fine, but if the main contribution of the paper involves human subjects, then as much detail as possible should be included in the main paper. 
        \item According to the NeurIPS Code of Ethics, workers involved in data collection, curation, or other labor should be paid at least the minimum wage in the country of the data collector. 
    \end{itemize}

\item {\bf Institutional review board (IRB) approvals or equivalent for research with human subjects}
    \item[] Question: Does the paper describe potential risks incurred by study participants, whether such risks were disclosed to the subjects, and whether Institutional Review Board (IRB) approvals (or an equivalent approval/review based on the requirements of your country or institution) were obtained?
    \item[] Answer: \answerNo{}
    \item[] Justification: Section~\ref{sec:ethics} states only that no personally identifying participant information is released. The study consists of domain experts rating blinded molecular proposals and carries no risk.
    \item[] Guidelines:
    \begin{itemize}
        \item The answer \answerNA{} means that the paper does not involve crowdsourcing nor research with human subjects.
        \item Depending on the country in which research is conducted, IRB approval (or equivalent) may be required for any human subjects research. If you obtained IRB approval, you should clearly state this in the paper. 
        \item We recognize that the procedures for this may vary significantly between institutions and locations, and we expect authors to adhere to the NeurIPS Code of Ethics and the guidelines for their institution. 
        \item For initial submissions, do not include any information that would break anonymity (if applicable), such as the institution conducting the review.
    \end{itemize}

\item {\bf Declaration of LLM usage}
    \item[] Question: Does the paper describe the usage of LLMs if it is an important, original, or non-standard component of the core methods in this research? Note that if the LLM is used only for writing, editing, or formatting purposes and does \emph{not} impact the core methodology, scientific rigor, or originality of the research, declaration is not required.
    %this research? 
    \item[] Answer: \answerYes{}
    \item[] Justification: An LLM is the core of the contribution, not a writing aid. Section~\ref{sec:system} identifies Claude Sonnet 4.6 as the configured reasoning backend that authors and revises objectives, invokes tools and selects candidates, and one of the two evaluated generation routes has the LLM propose molecules directly. Section~\ref{sec:system-prompt} describes the system prompt, states that provider-side sampling defaults were used with no fixed seed, and restricts the analysis to observable messages, tool requests, tool results and configuration payloads.
    \item[] Guidelines:
    \begin{itemize}
        \item The answer \answerNA{} means that the core method development in this research does not involve LLMs as any important, original, or non-standard components.
        \item Please refer to our LLM policy in the NeurIPS handbook for what should or should not be described.
    \end{itemize}

\end{enumerate}

\end{document}